\documentclass[a4paper,fleqn]{cas-dc}

\usepackage[numbers]{natbib}

\usepackage{booktabs}
\usepackage{multirow}
\usepackage{siunitx}
\usepackage{graphicx}
\usepackage{amsmath}
\usepackage{amssymb}
\usepackage{float}  
\usepackage{placeins}
\usepackage{capt-of}
\usepackage{expl3}
\ExplSyntaxOn
\makeatletter
\cs_set:Npn \__first_footerline:
  {
    \group_begin:
      \small\sffamily
      \ifnum\theblind>0\relax
      \else
        \__short_authors:
      \fi
    \group_end:
  }
\makeatother
\ExplSyntaxOff

\begin{document}
\let\printorcid\relax
\let\WriteBookmarks\relax

\setlength{\floatsep}{8pt plus 2pt minus 2pt}
\setlength{\textfloatsep}{10pt plus 2pt minus 4pt}
\setlength{\intextsep}{8pt plus 2pt minus 2pt}

\shorttitle{LASSA Architecture-Based Autonomous Fault-Tolerant Control of UUVs}

\title[mode=title]{LASSA Architecture-Based Autonomous Fault-Tolerant Control of Unmanned Underwater Vehicles}

\author[1]{Hong Chen}
\cormark[1]
\ead{chenhong22001@163.com}

\author[1,2]{Zixiang Tang}

\author[1]{Yuanbao Chen}

\author[1]{Yu Liu}

\affiliation[1]{organization={},
  addressline={Wuhan Second Ship Design and Research Institute},
  city={Wuhan},
  postcode={430060},
  country={China}}

  \affiliation[2]{organization={},
  addressline={School of Aeronautic Science and Engineering, Beihang University},
  city={Beijing},
  postcode={100191},
  country={China}}

\cortext[1]{Hong Chen.}

\begin{abstract}
Unmanned underwater vehicles (UUVs) operate persistently in communication-constrained environments, thus requiring high-level autonomous fault-tolerant control under faulty operating conditions. Existing approaches rely heavily on predefined hard-coded rules and struggle to achieve effective fault-tolerant control against unforeseen faults. Although large language models (LLMs) possess powerful cognitive and reasoning capabilities, their inherent hallucinations remain a major obstacle to their application in UUV control systems. This paper proposes an intelligent control method based on the LASSA (LLM-based Agent with Solver, Sensor and Actuator) architecture. Within this architecture, an LLM identifies unknown faults and accomplishes task replanning via autonomous reasoning without hard-coded rules; the intelligent agent undertakes perception, scheduling and decision evaluation; the solver verifies physical boundary feasibility constraints prior to command transmission to the actuators. This architecture suppresses physically infeasible LLM hallucinations and ensures interpretable, verifiable decision-making. Moreover, it enables fast–slow dual closed-loop collaborative control, where the slow loop undertakes high-level dynamic decision-making and the fast loop guarantees high-frequency real-time control, simultaneously balancing decision intelligence and control timeliness. Lake experiments under normal and lower-rudder-fault conditions show that the framework detects trajectory tracking abnormalities, replans the route by adjusting the turning radius from 4m to 12m and reducing speed from 2kn to 1kn, passes all three solver constraints on the first invocation, and guides the UUV to complete the full mission; under normal conditions no false fault alarms are raised throughout the run. Additional simulation experiments under abnormal scenarios such as cross-current disturbances and DVL sensor faults further demonstrate the robustness of the framework.
\end{abstract}



\maketitle

\section{Introduction}
Unmanned underwater vehicles (UUVs) have been widely used in marine observation, underwater inspection, environmental monitoring, and special operation support because of their compact structure, low deployment cost, and strong environmental adaptability \citep{ma2020pathreview, zeng2018pathreview}. As mission profiles extend to deeper waters, longer endurance operations, and increasingly remote deployment areas, UUVs are required to operate in environments where the underwater acoustic communication channel imposes severe constraints on bandwidth, propagation delay, and link availability \citep{liu2026cbba}. In such conditions, real-time teleoperation or shore-based intervention is impractical: the vehicle must rely on autonomous decision-making capabilities to handle unexpected events without human assistance. Among these, actuator faults represented by rudder failures are particularly critical, as they directly degrade heading control, trajectory tracking, and return safety. In addition, navigation equipment malfunctions and current disturbances can also severely impair the trajectory control performance of UUVs \citep{tian2022thruster, zhang2025multimodeftc}. Once the UUV can no longer follow the preset route and no remote operator is reachable in time, a delayed or inappropriate autonomous response may cause mission interruption, navigation instability, or even vehicle loss. Therefore, an architecture with high autonomy and precise control capability should be established to organically integrate state perception, fault analysis, mission replanning, solver-based physical verification, and actuation control into a complete closed-loop decision-making process, so as to guarantee fully autonomous operation of the system without any external input.

Existing studies on UUV fault handling mainly focus on fault diagnosis, fault-tolerant control, control allocation, and trajectory replanning \citep{tian2022thruster, chen2025ftmpc, wang2022finite, zhang2023predefined, wang2023metaself}. These methods have provided effective support for stabilizing the vehicle after actuator degradation or disturbance. However, most existing approaches are centered on low-level controller design or predefined rule-based adjustment, and they are typically optimized for relatively clear fault forms and known dynamic constraints \citep{fu2024takagi, yang2022apismc}. More critically, they implicitly assume that a human operator or a continuously available remote system remains in the loop to provide supervisory guidance whenever a fault-induced deviation exceeds the scope of the pre-designed controller. In the scenario where the communication link is constrained or severed and autonomous fault recovery is most urgently needed, there is no fallback cognitive mechanism to reassess the situation, reformulate the mission, and generate an adaptive response. Therefore, a time-division dual-loop control architecture with fast and slow loops should be constructed to organically integrate state perception, fault analysis, mission replanning, solver-based physical verification, and actuation control into a complete and coherent closed-loop decision-making process. The fast loop achieves real-time high-precision low-level control, while the slow loop performs dynamic optimization and adjustment of the control strategy, enabling the system to operate with high autonomy without any external input.

Recent advances in large language models (LLMs) and intelligent agent systems provide a new possibility for addressing this problem \citep{vroon2023chatgptrobotics, huang2024multilayerllm}. Unlike conventional fault-tolerant controllers that require the designer to enumerate all anticipated fault types and hard-code a corresponding response for each, an LLM reasons over the current vehicle state, fault description, and mission context to generate an adaptive replanning strategy without any pre-programmed rule set. This reasoning capability confers several properties that conventional methods cannot provide. First, the LLM operates autonomously: it completes the full fault-response cycle comprising perception, causal reasoning, trajectory replanning, and execution without requiring any communication with a remote operator. Second, it generalises to fault configurations that were never explicitly anticipated during design, because its response is derived from reasoning rather than table lookup. Third, it eliminates rule-combination explosion: as the number of possible fault types and environmental conditions grows, a rule-based system requires exponentially more hand-crafted cases, whereas the LLM adapts through context. These properties are not unique to underwater vehicles; the same architecture is applicable wherever communication constraints force fully autonomous fault recovery, including ground robots, manipulators, unmanned aerial vehicles, and autonomous driving systems under signal interference \citep{simon2024neurosymbolic, bhat2025brainbodyllm}. Nevertheless, directly deploying an LLM in the UUV control loop without additional safeguards introduces a critical risk: a strategy generated from semantic reasoning may be logically coherent yet physically infeasible given the vehicle's degraded actuator authority and environmental boundaries \citep{yang2023oceanchat, xu2025lyapunovauv}. Therefore, the core challenge is not simply incorporating an LLM into the control architecture, but grounding its outputs through a deterministic physical verification layer before they reach the actuators, thus forming a complete and reliable autonomous decision loop. Meanwhile, directly introducing LLM reasoning into the real-time control loop incurs non-negligible latency, making it difficult to satisfy high-precision control requirements. Therefore, the proposed architecture must strike a balance between autonomous capability and control precision.

To address the above issues, this paper proposes an intelligent control method for autonomous navigation of UUVs under fault conditions based on the LASSA (LLM-based Agent with Solver, Sensor and Actuator) architecture, as illustrated in Figure 1. This framework integrates the cognitive reasoning capability of LLMs, the perception-planning-scheduling capability of the agent, and the physical constraint validation capability supported by the solver. Within this architecture, the agent acquires real-time operational state data from onboard sensors to autonomously detect system anomalies. With the support of the LLM, the agent performs fault mechanism analysis and adaptively adjusts and optimizes control strategies. The LLM conducts logical reasoning based on the real-time UUV status and mission scenarios, and generates replanned control strategies and corresponding parameters without human intervention. The solver conducts physical constraint validation on the generated control strategies. Only when all constraints are satisfied can compliant commands be transmitted to the actuator layer for execution. Meanwhile, LASSA is formulated as a time-division fast-slow dual-loop control architecture. The fast loop realizes real-time high-precision low-level control, while the slow loop performs dynamic optimization and adjustment of control strategies. This method constructs a collaborative dual-loop system that integrates state perception, LLM reasoning and dynamic strategy generation, solver-based physical verification, and real-time low-level actuator control, enabling independent, reliable and fully autonomous operation in communication-constrained environments. The main contributions of this study are as follows.

First, a hierarchical intelligent control framework based on the LASSA architecture is proposed for fault-state UUV autonomous navigation under communication-constrained deployment conditions, enabling the vehicle to independently complete the full fault-response cycle spanning state perception, LLM-based cognitive reasoning, solver-based physical verification, and actuator execution, without relying on real-time remote intervention and without hard-coding any predefined rule set for fault handling. Unlike conventional fault-tolerant control methods that require explicit enumeration of fault types and corresponding response rules \citep{sinha2024cep, zhang2025multimodeftc}, the LLM component autonomously reasons over the current vehicle state, mission context, and fault description to generate adaptive replanning strategies, thereby avoiding rule-combination explosion in complex or compound fault scenarios. To the best of the authors' knowledge, this represents the first integration of LLM-driven fault-aware replanning with a deterministic solver verification layer specifically designed for rudder-degraded underwater navigation.

Second, a solver-based physical verification module is designed that certifies LLM-generated trajectory strategies against three physical constraints, namely boundary feasibility, speed limits, and a fault-aware minimum turning radius derived from the degraded actuator configuration, before actuator dispatch. This constraint-grounding mechanism suppresses physically infeasible LLM hallucinations from reaching the vehicle, an issue that existing LLM-for-robotics frameworks have left largely unaddressed. Furthermore, because every replanning decision passes through explicit, inspectable solver checks, the framework provides interpretable decision grounding: the reasoning path from fault detection through strategy generation to constraint certification is transparent and auditable, a property absent from end-to-end learned controllers.

Third, the proposed LASSA architecture adopts a time-sharing closed-loop configuration with fast and slow dual loops. The fast loop delivers real-time and high-precision responses for low-level actuation control, whereas the slow loop performs dynamic decision-making and optimization tuning based on LLM reasoning and physical solver verification. This design substantially improves the autonomous capability of the control system, while fully meeting the real-time and high-precision demands of low-level control.

Fourth, lake field trials are carried out under normal operating conditions and incipient rudder actuator faults to validate the end-to-end practicability of the proposed framework. Experimental results show that the LASSA architecture can monitor and analyze UUV rudder actuator faults. Leveraging the autonomous reasoning capability of the LLM, the system dynamically generates feasible replanning paths, which are transmitted to the actuators for implementation after passing physical verification via the solver. The results confirm that LASSA’s dynamic replanning and low-level real-time control capabilities well satisfy the fault-tolerant control demands of UUVs subjected to abnormal rudder actuator conditions.
\begin{figure*}[!ht]
  \centering
  \includegraphics[width=0.9\linewidth]{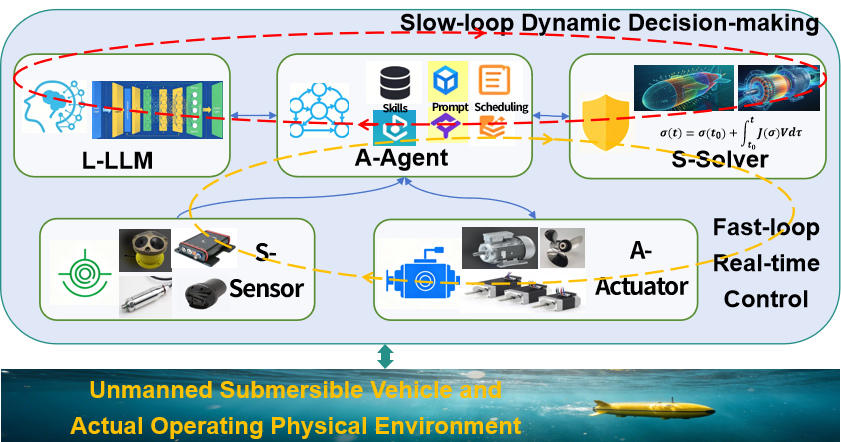}
  \caption{\textbf{Fault-Tolerant Control of UUV Based on LASSA Architecture. The LASSA architecture incorporates cognitive reasoning of large language models, perception-planning-scheduling of the agent, and physical constraint verification of the solver. The agent acquires operational data via onboard sensors to autonomously detect fault anomalies, performs fault mechanism analysis based on the LLM, and adaptively regulates and optimizes control strategies. After passing physical constraint verification by the solver, qualified control commands are transmitted to the actuator layer for implementation. The architecture adopts a time-division fast-slow dual closed-loop framework. The fast loop ensures high-precision real-time low-level control, while the slow loop achieves dynamic optimization of control strategies, enabling unmanned underwater vehicles to maintain highly autonomous control in communication-constrained scenarios.}}
  \label{fig:LASSA Architecture}
\end{figure*}

The remainder of this paper is organized as follows. Section~2 reviews the most relevant studies on UUV fault-tolerant control, path planning, and LLM-enabled robotic autonomy. Section~3 presents the LASSA architecture and the detailed design of each module. Section~4 describes the experimental design and implementation. Section~5 reports the experimental results and analysis, followed by conclusions and directions for future work.

    \section{Related Work}
Research closely related to this paper can be grouped into three directions: UUV fault-tolerant control and fault diagnosis, UUV path planning and online replanning, and LLM-enabled robotic decision-making. Underlying all three directions is a deployment reality that has received comparatively little explicit treatment in the literature: acoustic underwater communication channels are characterised by limited bandwidth (typically on the order of kilobits per second), high and variable propagation delay, and intermittent link availability in complex bathymetric environments \citep{liu2026cbba}. These constraints make real-time shore-based supervision impractical for long-endurance or remote missions, elevating the importance of fully autonomous onboard decision-making, particularly in fault scenarios where the vehicle cannot defer to a remote operator.

In the first direction, recent studies have investigated fault-tolerant control strategies for autonomous underwater vehicles under actuator degradation, unknown disturbances, saturation, and fault reconstruction constraints, including thruster-fault diagnosis and control co-design, model predictive fault-tolerant control, predefined-time or finite-time fault handling, and fuzzy or sliding-mode-based formulations \citep{tian2022thruster, chen2025ftmpc, wang2022finite, zhang2023predefined, fu2024takagi, yang2022apismc, tan2022fuzzyiter}. More recent work has extended these approaches to multi-mode thruster failure under complex environmental conditions and to deep-reinforcement-learning-based rudder fault diagnosis for X-rudder AUVs \citep{zhang2025multimodeftc, li2025xrudderdqn}. In parallel, data-driven diagnosis methods have been introduced for underwater vehicle health monitoring, including multi-scale attention-based diagnosis, weak fault detection, deep-learning-based diagnosis under missing data, and propeller fault diagnosis using particle-filter-based approaches \citep{wang2023metaself, wang2020tristable, wang2019deeplearnfd, liu2023missingdata, vukic2018rankpf, yan2018gprpf}. While these studies provide valuable support for fault identification and low-level stabilization, they primarily address diagnosis or controller design at the individual module level and do not provide a mechanism for translating detected faults into high-level trajectory replanning decisions. More critically, they do not address the scenario in which the vehicle must complete this translation autonomously, without any possibility of remote operator involvement.

The second direction concerns path planning, adaptive navigation, and online replanning for autonomous underwater vehicles. Review studies have systematically analyzed optimization-based, biologically inspired, and learning-based planning methods \citep{ma2020pathreview, zeng2018pathreview, yan2021researchprogress, yao2016survey3d, galceran2021coveragecpp}. More targeted work has addressed multi-AUV cooperative planning under realistic communication degradation, coverage path planning with real-time replanning, bioinspired neural-network planning, reinforcement-learning-based navigation, and path generation in unknown underwater canyons \citep{abbasi2023energyefficient, englot2014coverage, fan2024noisydueling, cheng2022sac, zhu2017bioinspired, li2021underwatercanyons}. Although these methods significantly improve planning quality and online route generation, they presuppose that replanning can be triggered by well-defined external signals or that the planning system operates in isolation from the fault-detection layer. None of them closes the loop between fault-induced state deviation, autonomous cognitive replanning, and physical feasibility verification in a manner that remains fully operable without external communication.

The third direction involves LLMs and agent architectures for robotic systems. Broad studies have demonstrated that LLMs can support robot task planning, instruction interpretation, action sequencing, and high-level decision coordination \citep{vroon2023chatgptrobotics, huang2024multilayerllm, simon2024neurosymbolic, boiko2023chemrobotics, li2025vlagrasp, mower2026rosllm}. More recent embodied frameworks such as ELLMER further couple LLMs with retrieval-augmented generation and sensorimotor feedback to complete long-horizon tasks in unpredictable settings \citep{monwilliams2025ellmer}. A particularly relevant line of work addresses LLM-based online replanning: RePLan \citep{skreta2024replan} uses perception and language models to adapt robot actions when an initial plan fails due to unforeseen obstacles, while BrainBody-LLM \citep{bhat2025brainbodyllm} couples a hierarchical two-LLM planner with closed-loop execution feedback for real-time replanning. Brain-inspired embodied agents that explicitly partition cognition and dynamics, such as EvoAgent \citep{gao2025evoagent}, which couples a multimodal LLM with a world model through dynamic communication slots, further illustrate the broader trend toward dual-system architectures. These works demonstrate the viability of LLM-driven replanning but operate in terrestrial manipulation settings with persistent cloud connectivity and do not address physical constraint certification of replanned trajectories. In the underwater domain, recent work has explored LLMs for AUV attitude stability and adaptive control under hydrodynamic disturbances \citep{xu2025lyapunovauv, cai2025nevertooprim}, and language-driven command generation for underwater, subsea, and surface vehicles has been demonstrated by OceanChat, AquaChat, Word2Wave, and AI~Captain \citep{yang2023oceanchat, lee2025aquachat, hou2024word2wave, christensen2025aicaptain}. However, these works predominantly rely on persistent connectivity between the vehicle and a remote or cloud-hosted LLM service, and they address motion control adaptation or natural-language task interaction rather than autonomous fault-state recovery. Ensuring that a generated trajectory is actually executable given the vehicle's degraded actuator authority and environmental boundary constraints remains unaddressed across all of them. None of these works addresses the compound scenario in which communication constraints eliminate the possibility of remote intervention while a complex actuator fault simultaneously requires the vehicle to autonomously complete the full cognitive cycle of abnormality detection, LLM-based trajectory replanning, solver-based physical constraint verification, and actuator execution within a single closed-loop architecture. This gap in combining communication-constrained deployment, complex rudder fault handling, and physically verified autonomous replanning directly motivates the LASSA framework proposed in this paper. To the best of the authors' knowledge, this is the first work to couple LLM-driven fault-aware trajectory replanning with a deterministic solver verification layer in a fully autonomous closed-loop control architecture designed for communication-constrained UUV deployment under actuator faults.

\section{Method}

As established in Section~1, fault-state UUV navigation under communication constraints presents a compound challenge: the vehicle must complete the full fault-response cycle of detection, reasoning, replanning, verification, and execution entirely onboard, without access to a remote operator or cloud-based planner. Conventional fault-tolerant control methods address this through pre-designed rule sets or model-based controllers that enumerate anticipated fault modes \citep{sinha2024cep, zhang2025multimodeftc}, but such approaches are inherently brittle: their response space is bounded by the designer's foresight and cannot adapt to fault configurations or environmental conditions that were not explicitly anticipated. The LASSA architecture addresses this limitation by substituting the fixed rule set with an LLM-based cognitive reasoning module that autonomously interprets the current fault state and mission context to generate adaptive replanning strategies, thereby eliminating rule-combination explosion and enabling generalisation to complex fault scenarios without additional programming. The solver layer then acts as a hard physical filter, certifying each LLM-generated strategy before it reaches the actuators and suppressing physically infeasible outputs that may arise from pure language-model reasoning. This separation of cognitive autonomy (LLM) from physical feasibility enforcement (solver) also confers interpretability: every replanning decision is grounded by explicit, inspectable constraint checks, making the framework's decision process transparent and auditable.

\subsection{Overview}

\begin{figure*}[!ht]
  \centering
  \includegraphics[width=0.9\linewidth]{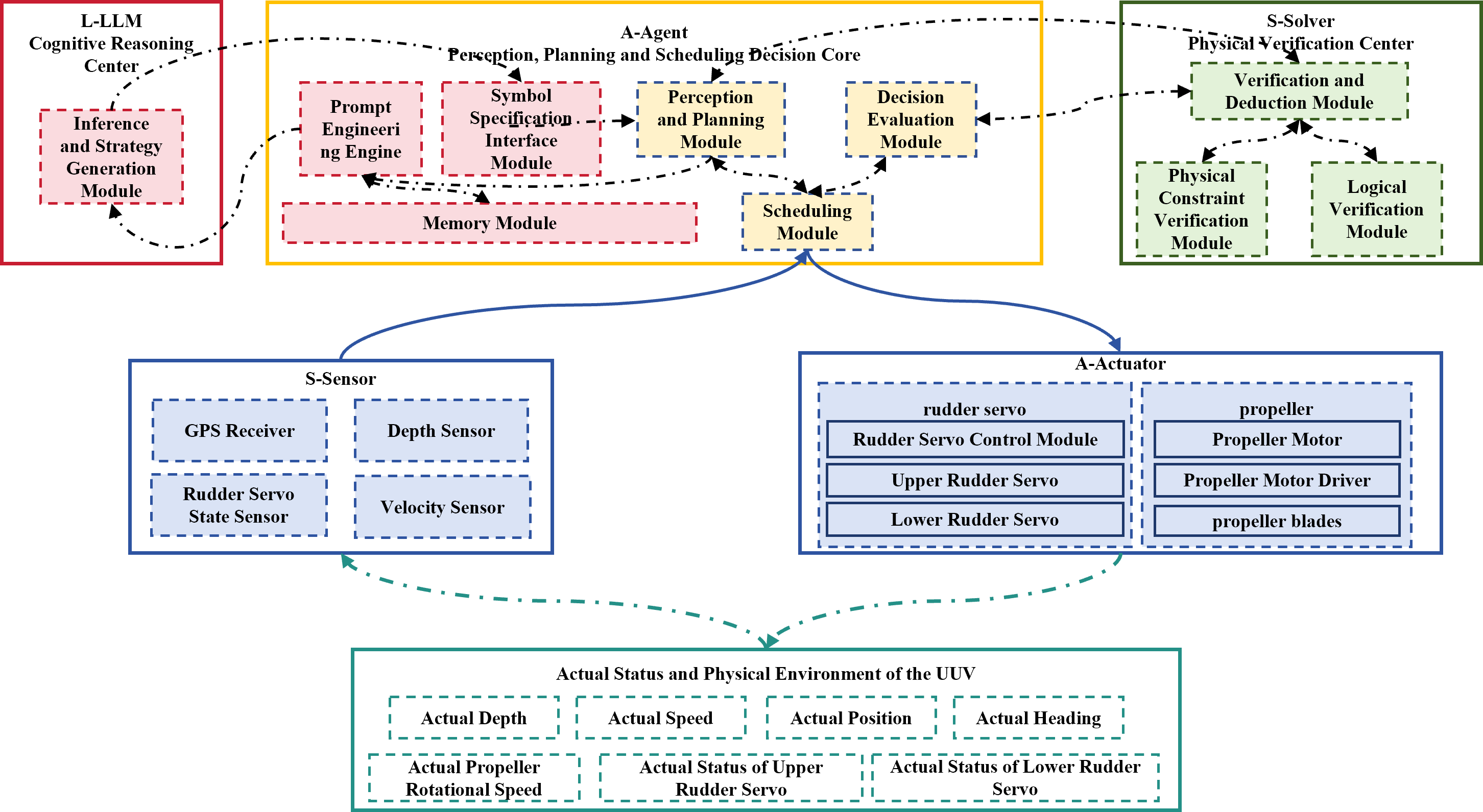}
  \caption{Principle diagram of LASSA-based intelligent control for UUV autonomous navigation}
  \label{fig:lassa_architecture}
\end{figure*}

The LASSA architecture is adopted in this study for UUV autonomous navigation control. As illustrated in Figure~\ref{fig:lassa_architecture}, the framework is organized into five tightly coupled layers: a large language model (LLM) for high-level cognitive reasoning, an intelligent agent for perception, planning, scheduling, and decision evaluation, a solver-based physical verification center for feasibility and constraint checking, a sensor layer for real-time state acquisition, and an actuator layer for command execution. During operation, data flow upward from the sensor layer through the agent to the LLM when needed, and commands flow downward from the decision evaluation module through the solver and back to the actuators. The remainder of this section describes each layer in turn, beginning with the kinematic model that underpins the solver and control laws, followed by the agent core, the LLM module, the solver, and finally the sensor and actuator layers.

The complete system state at time $t$ is defined as the vector
\begin{equation}
  \mathbf{s}(t) = \bigl[x(t),\; y(t),\; \psi(t),\; u(t),\; d(t),\; \delta_r(t)\bigr]^\top,
  \label{eq:state_vec}
\end{equation}
where $(x, y)$ denotes the horizontal position in a fixed navigation frame, $\psi$ is the heading angle, $u$ is the surge speed, $d$ is the diving depth, and $\delta_r \in \{0,1\}$ is a binary rudder-health indicator ($\delta_r = 0$ denotes a fault; $\delta_r = 1$ denotes healthy). The first five components are continuous and take values in $\mathbb{R}$; $\delta_r$ is treated as an appended discrete flag for notational convenience. Under normal operation, the agent executes a closed-loop navigation cycle driven by sensor measurements; upon detecting a navigation abnormality, the system escalates to the LLM layer for fault-aware trajectory replanning, passes the candidate strategy to the solver for physical verification, and finally delivers the verified command $\mathbf{C}(t)$ to the actuator layer. The overall closed-loop control mapping is written as
\begin{equation}
  \mathbf{C}(t) = \mathcal{K}\!\bigl(\hat{\mathbf{s}}(t),\; \mathbf{W}_\mathrm{act}(t),\; \hat{\alpha}(t)\bigr),
  \label{eq:control_mapping}
\end{equation}
where $\mathbf{W}_\mathrm{act}(t)$ is the \emph{active} waypoint sequence at time $t$, equal to $\mathbf{W}_\mathrm{ref}$ (the original plan) before any fault-triggered replanning and updated to $\mathbf{W}_\mathrm{new}$ (the LLM-replanned route) once a solver-verified replanning has been accepted; $\hat{\alpha}(t) \in \{0,1\}$ is the confirmed fault flag generated by the perception planning module (formally defined in \eqref{eq:confirmed_flag}), and $\hat{\mathbf{s}}(t)$ denotes the estimated vehicle state derived from the sensor measurements $\mathbf{z}_c(t)$ and $\delta_r^m(t)$.

\subsection{UUV Kinematic Model and State Representation}

To provide a physical basis for trajectory planning, abnormality detection, and solver verification, the horizontal-plane kinematics of the UUV are described by the three-degree-of-freedom model
\begin{equation}
  \dot{x} = u\cos\psi - v\sin\psi, \quad
  \dot{y} = u\sin\psi + v\cos\psi, \quad
  \dot{\psi} = r,
  \label{eq:kinematics}
\end{equation}
where $v$ is the lateral (sway) velocity and $r$ is the yaw rate. In the shallow-water, low-speed regime of the lake experiments, the sway component $v$ is relatively small and is treated as a bounded disturbance. The heading controller is a PD law acting on the estimated heading error:
\begin{equation}
  \delta(t) = K_p\,e_\psi(t) + K_d\,\dot{e}_\psi(t),
  \label{eq:pd_heading}
\end{equation}
where $e_\psi(t) = \psi_\mathrm{des}(t) - \hat{\psi}(t)$ is the heading error computed from the estimated heading $\hat{\psi}(t)$, $\psi_\mathrm{des}$ is the desired heading provided by the LOS guidance law (defined in \eqref{eq:los} in Section~\ref{sec:actuators}), and $K_p$, $K_d$ are the proportional and derivative gains, respectively. The rudder command $\delta(t)$ is saturated within $[-\delta_\mathrm{max},\,\delta_\mathrm{max}]$ to respect actuator limits.

The trajectory reference is defined as an ordered sequence of $N$ waypoints
\begin{equation}
  \mathbf{W}_\mathrm{ref} = \bigl\{w_k = (x_k^r,\, y_k^r)\bigr\}_{k=1}^{N},
  \label{eq:waypoints}
\end{equation}
with each inter-waypoint transition described by a circular arc of turning radius $R_k$ and a constant forward speed $u_k$. The UUV switches to the next waypoint $w_{k+1}$ when the cross-track distance to the current segment falls below a waypoint-acceptance radius $r_\mathrm{acc}$.

The cross-track error is defined with respect to the nearest point on the reference path. In terms of the true state it is
\begin{equation}
  e_p(t) = \sqrt{\bigl(x(t)-x_\mathrm{ref}(t)\bigr)^2 + \bigl(y(t)-y_\mathrm{ref}(t)\bigr)^2},
  \label{eq:crosstrack}
\end{equation}
where $(x_\mathrm{ref}(t), y_\mathrm{ref}(t))$ is the nearest point on the reference path at time $t$. In the implementation, the estimated position $(\hat{x}(t), \hat{y}(t))$ from $\hat{\mathbf{s}}(t)$ is substituted for $(x(t), y(t))$.

\subsection{Agent Core for Perception, Planning, and Scheduling}
\label{sec:agent_core}

The intelligent agent constitutes the computational core of the LASSA architecture, responsible for global scheduling, state perception, decision-making, and execution management. It consists of three collaborating modules: the perception planning module, the decision evaluation module, and the scheduling module.

\subsubsection{Perception Planning Module}

At each control cycle of duration $\Delta t$, the perception planning module receives the structured state message $\mathbf{m}(t)$ assembled by the scheduling module:
\begin{equation}
  \mathbf{m}(t) = \bigl\langle \hat{\mathbf{s}}(t),\; e_p(t),\; e_\psi(t),\; \delta_r^m(t),\; t \bigr\rangle,
  \label{eq:state_message}
\end{equation}
where $\hat{\mathbf{s}}(t)$ is the estimated vehicle state derived from the sensor measurements $\mathbf{z}_c(t)$, and $e_p(t)$ and $e_\psi(t)$ are the cross-track and heading errors computed from $\hat{\mathbf{s}}(t)$ relative to the current reference waypoint. The module computes the fault flag by evaluating three conditions:
\begin{equation}
  \alpha(t) =
  \begin{cases}
    1, & \text{if } e_p(t) > \varepsilon_p \;\text{ or }\; |e_\psi(t)| > \varepsilon_\psi \;\text{ or }\; \delta_r^m(t) = 0, \\
    0, & \text{otherwise},
  \end{cases}
  \label{eq:fault_flag}
\end{equation}
where $\varepsilon_p$ (m) and $\varepsilon_\psi$ (rad) are the position and heading deviation thresholds, respectively, and $\delta_r^m(t) = 0$ directly indicates a hardware rudder fault reported by the rudder-state sensor. To avoid spurious triggering caused by transient sensor noise, the flag is confirmed only when the condition in \eqref{eq:fault_flag} persists for a consecutive window of $n_w$ cycles:
\begin{equation}
  \hat{\alpha}(t) = \mathbf{1}\!\left[\sum_{i=0}^{n_w-1}\alpha(t - i\Delta t) \geq n_w\right].
  \label{eq:confirmed_flag}
\end{equation}
Here the hat in $\hat{\alpha}$ denotes \emph{confirmation} (persistent detection over $n_w$ cycles), not estimation in the statistical sense; $\hat{\alpha}(t) = 1$ is the signal that activates the LLM reasoning pipeline.
When $\hat{\alpha}(t) = 0$, the module passes the current trajectory parameters directly to the decision evaluation module for routine execution. When $\hat{\alpha}(t) = 1$, it assembles the context package $\mathcal{I}(t) = \langle \hat{\mathbf{s}}(t),\; \mathbf{W}_\mathrm{ref},\; \mathcal{H}(t) \rangle$ and triggers the LLM reasoning pipeline, where $\mathcal{H}(t)$ denotes the historical context retrieved from the memory module.

\subsubsection{Decision Evaluation Module}

The decision evaluation module operates in two modes depending on whether the LLM has been invoked. Under normal operation ($\hat{\alpha}(t) = 0$), no LLM call is made and the module directly issues the trajectory-tracking command:
\begin{equation}
  \mathbf{C}(t) = \mathcal{C}_\mathrm{track}\!\bigl(\hat{\mathbf{s}}(t),\; w_k^\mathrm{act}\bigr), \quad \hat{\alpha}(t) = 0,
  \label{eq:normal_cmd}
\end{equation}
where $\mathcal{C}_\mathrm{track}$ encodes the LOS heading law \eqref{eq:los} and the segment plan speed, and $w_k^\mathrm{act}$ is the currently active waypoint, drawn from $\mathbf{W}_\mathrm{ref}$ before any fault-triggered replanning or from $\mathbf{W}_\mathrm{new}$ (as defined in \eqref{eq:llm_output}) once a replanned route has been successfully verified and activated. Under fault operation ($\hat{\alpha}(t) = 1$), after the LLM generates a candidate strategy $\Theta(t)$ (defined in Section~\ref{sec:llm}), the module submits it to the physical solver (Section~\ref{sec:solver}) and obtains a binary verification result
\begin{equation}
  v = \mathcal{V}\!\bigl(\Theta(t)\bigr) \in \{\textsc{pass},\, \textsc{fail}(\mathcal{E})\},
  \label{eq:verify_result}
\end{equation}
where $\mathcal{E}$ is the set of violated constraint labels returned on failure. The module then applies the execution decision rule:
\begin{equation}
  \mathbf{C}(t) =
  \begin{cases}
    \mathcal{H}_\mathrm{sym}\!\bigl(\Theta(t)\bigr), & \text{if } v = \textsc{pass}, \\
    \mathcal{K}_\mathrm{hold}(\hat{\mathbf{s}}(t)), & \text{if } v = \textsc{fail}(\mathcal{E}) \text{ and } n_\mathrm{retry} \geq n_\mathrm{max},
  \end{cases}
  \label{eq:decision_rule}
\end{equation}
where $\mathcal{H}_\mathrm{sym}$ is the symbolic interfacing function defined in Section~\ref{sec:llm}, $\mathcal{K}_\mathrm{hold}$ is a conservative hold-heading fallback controller that maintains the last valid heading and reduces speed to $u_\mathrm{min}$, and $n_\mathrm{retry}$ counts the number of LLM regeneration attempts against the maximum allowed $n_\mathrm{max}$. If $v = \textsc{fail}(\mathcal{E})$ and $n_\mathrm{retry} < n_\mathrm{max}$, the failure label set $\mathcal{E}$ is appended to the prompt and the LLM is invoked again.

\subsubsection{Scheduling Module}

The scheduling module manages the timing and data routing among all subsystems. At each cycle it reads the raw measurements $\mathbf{z}_c(t)$ and $\delta_r^m(t)$ from the sensor layer, forms the estimated state $\hat{\mathbf{s}}(t)$, computes the navigation errors $e_p(t)$ and $e_\psi(t)$ against the active waypoint, assembles the message $\mathbf{m}(t)$ defined in \eqref{eq:state_message}, and dispatches it to the perception planning module. Conversely, it receives the verified command list $\mathbf{C}(t)$ from the decision evaluation module and transmits each command to the actuator layer via the CAN bus. The inter-module message format is defined as a typed tuple
\begin{equation}
  \mathrm{msg} = \bigl(t_\mathrm{stamp},\; \mathrm{src},\; \mathrm{dst},\; \mathrm{type},\; \mathbf{payload}\bigr),
  \label{eq:message_format}
\end{equation}
where $t_\mathrm{stamp}$ is the UNIX timestamp, \texttt{src} and \texttt{dst} identify the sending and receiving modules, \texttt{type} specifies whether the payload carries state data, a planning request, a strategy, a verification result, or a control command, and $\mathbf{payload}$ contains the corresponding data structure.

\subsection{Cognitive Reasoning Based on a Large Language Model}
\label{sec:llm}

The LLM module serves as the high-level cognitive reasoning engine within the LASSA framework. It is invoked in two situations: (i) at mission start, to generate the initial trajectory plan from the mission specification (with $\hat{\alpha}(t) = 0$ and an empty fault context); and (ii) during runtime, when the confirmed fault flag $\hat{\alpha}(t) = 1$ is raised, to generate physically grounded trajectory replanning strategies. Once the LASSA control loop enters routine execution under $\hat{\alpha}(t) = 0$, no further LLM call is made until a fault is confirmed (cf.\ Section~\ref{sec:agent_core}). The module comprises four functional components: a prompt engine, a reasoning and strategy generation unit, a symbolic standard interfacing unit, and a memory module.

\subsubsection{Prompt Engine}

Given the context package $\mathcal{I}(t)$, the prompt engine constructs a structured natural-language prompt
\begin{equation}
  \mathcal{P}(t) = \mathcal{F}_\mathrm{prompt}\!\bigl(\hat{\mathbf{s}}(t),\; \mathcal{T},\; \mathcal{H}(t),\; \hat{\alpha}(t),\; \mathcal{E}\bigr),
  \label{eq:prompt_fn}
\end{equation}
where $\mathcal{T}$ is the mission description (target waypoints, mission priority, and safety requirements) and $\mathcal{E}$ is the constraint-violation feedback from the solver on retry attempts. The prompt is structured into four fixed sections: (i) \textit{System Role}, which defines the LLM's role as a UUV navigation decision agent; (ii) \textit{Vehicle State}, which encodes the estimated state $\hat{\mathbf{s}}(t)$ in tabular form; (iii) \textit{Task and Constraints}, which specifies the mission objective and the physical bounds imposed by the solver; and (iv) \textit{Abnormality Description}, which translates $\hat{\alpha}(t)$ and $\mathcal{E}$ into natural-language fault context. This structured template ensures deterministic output formatting and reduces the risk of physically infeasible hallucinations. On the first invocation $\mathcal{E} = \emptyset$ and section~(iv) describes only the detected navigation deviation; on retry invocations, $\mathcal{E}$ is populated with the names of the failed solver constraints, giving the LLM explicit corrective guidance.

\subsubsection{Reasoning and Strategy Generation}

Upon receiving $\mathcal{P}(t)$, the LLM performs chain-of-thought reasoning and generates a candidate replanning strategy expressed as a structured parameter set:
\begin{equation}
  \Theta(t) = \mathcal{G}_\mathrm{LLM}\!\bigl(\mathcal{P}(t)\bigr)
            = \bigl\{R_\mathrm{new},\; u_\mathrm{new},\; \mathbf{W}_\mathrm{new},\; \psi_\mathrm{ret}\bigr\},
  \label{eq:llm_output}
\end{equation}
where $R_\mathrm{new}$ is the proposed turning radius (m), $u_\mathrm{new}$ is the proposed navigation speed (kn), $\mathbf{W}_\mathrm{new} = \{w_1^\prime, \ldots, w_{N^\prime}^\prime\}$ is the replanned waypoint sequence, and $\psi_\mathrm{ret}$ is the desired heading at the return waypoint. The reasoning process considers the current fault type, the remaining mission distance, the lake boundary geometry, and the reduced actuator authority under the fault condition.

\subsubsection{Symbolic Standard Interfacing}

Because the LLM output $\Theta(t)$ contains structured numeric parameters (turning radius, speed, waypoint coordinates, return heading) embedded in natural-language text, the symbolic interfacing unit parses these values and assembles a strictly typed command list that can be dispatched directly to hardware:
\begin{equation}
  \mathbf{C}(t) = \mathcal{H}_\mathrm{sym}\!\bigl(\Theta(t)\bigr)
               = \bigl\{(c_k,\, p_k)\bigr\}_{k=1}^{K},
  \label{eq:symbolic_if}
\end{equation}
where $c_k \in \{\texttt{SET\_SPEED},\; \texttt{SET\_RADIUS},\; \texttt{ADD\_WAYPOINT},\; \texttt{SET\_HEADING}\}$ is a command token and $p_k$ is the corresponding numeric parameter. The conversion enforces unit consistency (converting natural-language units to SI) and clips values to pre-defined safe ranges; a parse error is raised if any mandatory field is absent. Note that $\mathcal{H}_\mathrm{sym}$ is applied \emph{after} the solver has returned a \textsc{pass} verdict on $\Theta(t)$ (see Section~\ref{sec:solver} and \eqref{eq:decision_rule}): the solver operates on the structured parameters in $\Theta(t)$ directly, and the symbolic conversion to $\mathbf{C}(t)$ is only performed once physical feasibility has been confirmed. The resulting command list $\mathbf{C}(t)$ is then forwarded to the scheduling module for actuator dispatch.

\subsubsection{Memory Module}

The memory module maintains two independent stores. The short-term context buffer $\mathcal{H}_\mathrm{ST}$ retains the last $M_\mathrm{ST}$ prompt–response pairs within the current mission session to enable coherent multi-turn reasoning:
\begin{equation}
  \mathcal{H}_\mathrm{ST}(t) = \bigl\{(\mathcal{P}(t_i),\; \Theta(t_i))\bigr\}_{i = \max(1,\, n-M_\mathrm{ST})}^{n},
  \label{eq:short_term}
\end{equation}
where $n$ is the index of the current reasoning invocation. The long-term knowledge base $\mathcal{H}_\mathrm{LT}$ stores mission summaries, known fault patterns, and previously verified recovery strategies retrieved by semantic similarity search. The combined context injected into the prompt engine is
\begin{equation}
  \mathcal{H}(t) = \mathcal{H}_\mathrm{ST}(t) \,\cup\, \mathcal{H}_\mathrm{LT}(t),
  \label{eq:memory_combined}
\end{equation}
where $\mathcal{H}_\mathrm{LT}(t)$ denotes the top-$K$ entries retrieved from the long-term store by semantic similarity to the current prompt $\mathcal{P}(t)$. Injecting $\mathcal{H}(t)$ into $\mathcal{P}(t)$ improves the accuracy and coherence of reasoning across repeated fault events within the same or subsequent missions.

\subsection{Solver-Based Physical Verification Center}
\label{sec:solver}

The solver acts as a safety gateway between the LLM-generated strategy $\Theta(t)$ and the actuator layer. Its purpose is to certify that the proposed trajectory is physically achievable given the UUV's kinematic limits, the environmental boundary of the test lake, and the degraded actuator authority caused by the rudder fault. The verification is decomposed into three sequential checks; the strategy passes only if all three are satisfied.

\subsubsection{Boundary Feasibility Check}

The navigable water area is represented as a convex polygon $\mathcal{W}$ with $N_b$ vertices $\{(x_i^b, y_i^b)\}_{i=1}^{N_b}$ in the same navigation frame as \eqref{eq:kinematics}. For each planned waypoint $w_k^\prime \in \mathbf{W}_\mathrm{new}$, the signed distance to the nearest boundary edge is computed as
\begin{equation}
  d_b(w_k^\prime) = \min_{i=1}^{N_b} \mathrm{dist}\!\bigl(w_k^\prime,\; e_i\bigr),
  \label{eq:boundary_dist}
\end{equation}
where $e_i$ is the $i$-th boundary edge. The boundary constraint requires
\begin{equation}
  d_b(w_k^\prime) \geq d_\mathrm{safe}, \quad \forall\, k = 1,\ldots, N^\prime,
  \label{eq:boundary_check}
\end{equation}
with safety margin $d_\mathrm{safe}$ (m). For each circular arc turn centered at $\mathbf{o}_k = (o_k^x, o_k^y)$ with radius $R_k$, the arc is sampled at angular resolution $\Delta\theta$ and every sample point is tested against \eqref{eq:boundary_check}:
\begin{equation}
  d_b\!\bigl(\mathbf{o}_k + R_k[\cos\theta,\, \sin\theta]^\top\bigr) \geq d_\mathrm{safe},
  \quad \theta \in [\theta_\mathrm{start},\, \theta_\mathrm{end}].
  \label{eq:arc_boundary}
\end{equation}
If any sample violates \eqref{eq:boundary_check}, the boundary check fails and the specific violated waypoints are included in $\mathcal{E}$ for LLM feedback.

\subsubsection{Speed and Actuator Constraint Check}

The proposed navigation speed $u_\mathrm{new}$ must satisfy the vehicle's speed envelope:
\begin{equation}
  u_\mathrm{min} \leq u_\mathrm{new} \leq u_\mathrm{max},
  \label{eq:speed_bound}
\end{equation}
where $u_\mathrm{min}$ is the minimum speed for maintaining rudder effectiveness and $u_\mathrm{max}$ is the rated maximum speed. In addition, executing a circular turn of radius $R$ at speed $u$ requires a lateral acceleration
\begin{equation}
  a_c = \frac{u_\mathrm{new}^2}{R_\mathrm{new}},
  \label{eq:centripetal}
\end{equation}
which must remain within the maximum lateral force that the rudder can deliver:
\begin{equation}
  a_c \leq a_\mathrm{max} = \frac{F_{r,\mathrm{max}}}{m},
  \label{eq:accel_limit}
\end{equation}
where $F_{r,\mathrm{max}}$ is the maximum lateral rudder force and $m$ is the vehicle mass. Combining \eqref{eq:centripetal} and \eqref{eq:accel_limit}, the speed-constrained minimum turning radius is
\begin{equation}
  R_\mathrm{min}(u_\mathrm{new}) = \frac{u_\mathrm{new}^2}{a_\mathrm{max}}.
  \label{eq:rmin_speed}
\end{equation}
The speed constraint check passes if and only if $R_\mathrm{new} \geq R_\mathrm{min}(u_\mathrm{new})$.

\subsubsection{Fault-Aware Minimum Turning Radius Check}

Under normal operation with both upper and lower vertical rudders active, the geometric minimum turning radius at maximum rudder deflection $\delta_\mathrm{max}$ is approximated by
\begin{equation}
  R_\mathrm{min}^\mathrm{nom} = \frac{L}{2\sin\delta_\mathrm{max}},
  \label{eq:rmin_nominal}
\end{equation}
where $L$ is the vehicle length. This expression is a first-order geometric approximation derived from the turning-circle geometry of a rigid-body vehicle \citep{ma2020pathreview} and is used here as a conservative lower bound for solver validation rather than a precise hydrodynamic prediction. When the lower rudder is removed, the effective maximum deflection is reduced to $\delta_\mathrm{max}^\mathrm{eff} < \delta_\mathrm{max}$ owing to the asymmetric fin configuration and the consequent degradation in yaw authority. The fault-state minimum turning radius is therefore
\begin{equation}
  R_\mathrm{min}^\mathrm{fault} = \frac{L}{2\sin\delta_\mathrm{max}^\mathrm{eff}} > R_\mathrm{min}^\mathrm{nom}.
  \label{eq:rmin_fault}
\end{equation}
The value of $\delta_\mathrm{max}^\mathrm{eff}$ is estimated offline from a set of captive turning-circle trials conducted before the lake experiment. The turning-radius constraint check passes if
\begin{equation}
  R_\mathrm{new} \geq \max\!\bigl(R_\mathrm{min}^\mathrm{fault},\; R_\mathrm{min}(u_\mathrm{new})\bigr).
  \label{eq:rmin_combined}
\end{equation}

\subsubsection{Composite Verification Decision}

Denoting the three check results as $v_b$ (boundary), $v_s$ (speed), and $v_R$ (turning radius), the overall solver verdict is
\begin{equation}
  \mathcal{V}\!\bigl(\Theta(t)\bigr) =
  \begin{cases}
    \textsc{pass},              & \text{if } v_b \wedge v_s \wedge v_R = \text{true},  \\
    \textsc{fail}(\mathcal{E}), & \text{otherwise},
  \end{cases}
  \label{eq:solver_verdict}
\end{equation}
where $\mathcal{E} \subseteq \{\texttt{boundary},\, \texttt{speed},\, \texttt{radius}\}$ is the set of failed constraint labels, defined as
\begin{equation}
  \mathcal{E} = \bigl\{\texttt{boundary} \mid \neg v_b\bigr\}
              \cup \bigl\{\texttt{speed} \mid \neg v_s\bigr\}
              \cup \bigl\{\texttt{radius} \mid \neg v_R\bigr\}.
  \label{eq:error_set}
\end{equation}
This structured label set is returned to the decision evaluation module and appended to the LLM prompt on retry invocations to provide explicit corrective guidance.

\subsection{Sensors}

The sensor layer collects the UUV physical state in real time. The continuous-valued components of the measurement vector at time $t$ are
\begin{equation}
  \mathbf{z}_c(t) = \bigl[x^m(t),\; y^m(t),\; \psi^m(t),\; u^m(t),\; d^m(t)\bigr]^\top,
  \label{eq:meas_vec}
\end{equation}
comprising GPS position $(x^m, y^m)$, compass heading $\psi^m$, speed-log surge velocity $u^m$, and pressure-based depth $d^m$. These are modeled as noisy observations of the corresponding continuous states:
\begin{equation}
  \mathbf{z}_c(t) = \mathbf{s}_c(t) + \boldsymbol{\eta}(t),\quad
  \boldsymbol{\eta}(t) \sim \mathcal{N}(\mathbf{0},\;\boldsymbol{\Sigma}_\eta),
  \label{eq:meas_model}
\end{equation}
where $\mathbf{s}_c(t) = [x, y, \psi, u, d]^\top$ and $\boldsymbol{\Sigma}_\eta = \mathrm{diag}(\sigma_x^2, \sigma_y^2, \sigma_\psi^2, \sigma_u^2, \sigma_d^2)$ is the measurement noise covariance. The rudder-health flag $\delta_r^m(t) \in \{0,1\}$ is acquired separately from a dedicated fault-detection circuit on the CAN bus and is treated as a deterministic boolean signal; it is not subject to additive Gaussian noise. All signals are sampled at $f_s = 10$ Hz, encoded as CAN frames, and transmitted to the shore-based agent via a 4G/5G wireless link.

\subsection{Actuators and Control Execution}
\label{sec:actuators}

The actuator layer constitutes the final execution stage of the LASSA control loop. Upon receiving the verified command $\mathbf{C}(t)$ from the scheduling module, it decomposes the command list into individual device setpoints and drives each device accordingly.

The \textit{heading control} loop runs at $f_c = 20$ Hz and consists of two cascaded elements: the line-of-sight (LOS) guidance law that generates the desired heading, and the PD controller that tracks it. The LOS law computes:
\begin{equation}
  \psi_\mathrm{des}(t) = \mathrm{atan2}\!\bigl(y_{k}^r - \hat{y}(t),\; x_{k}^r - \hat{x}(t)\bigr),
  \label{eq:los}
\end{equation}
where $(x_k^r, y_k^r)$ is the current target waypoint and $(\hat{x}(t), \hat{y}(t))$ are the estimated horizontal position components from $\hat{\mathbf{s}}(t)$. The resulting $\psi_\mathrm{des}(t)$ is fed into the PD controller defined in \eqref{eq:pd_heading}, and the resulting rudder command $\delta(t)$ is mapped to a CAN setpoint after saturation clamping:
\begin{equation}
  \delta^\mathrm{cmd}(t) = \mathrm{clip}\!\bigl(\delta(t),\; -\delta_\mathrm{max},\; \delta_\mathrm{max}\bigr).
  \label{eq:rudder_cmd}
\end{equation}
The \textit{propulsion control} loop receives the currently commanded speed setpoint $u^\mathrm{cmd}$, which equals $u_k$ (the plan speed of the active waypoint segment) under normal operation or $u_\mathrm{new}$ (the LLM-replanned speed) after a fault is detected. The loop drives the brushless DC motor via a dedicated speed controller to maintain the target RPM within $\pm 5$ r/min. When a \texttt{SET\_SPEED} command is issued under fault conditions, the speed setpoint is additionally clipped to the fault-safe upper bound $u_\mathrm{max}^\mathrm{fault} \leq u_\mathrm{max}$ to limit the centripetal demand on the degraded rudder. The fault-safe bound is derived from the combined turning-radius constraint \eqref{eq:rmin_combined}:
\begin{equation}
  u_\mathrm{max}^\mathrm{fault} = \sqrt{a_\mathrm{max}\cdot R_\mathrm{min}^\mathrm{fault}}.
  \label{eq:umax_fault}
\end{equation}
All actuator commands are logged to the shore-based monitoring system with their associated timestamps to support post-experiment trajectory reconstruction and performance analysis.

\section{Experimental Design and Implementation}

\subsection{Experimental Objectives}

The lake experiments are designed to validate the proposed LASSA framework under two operating conditions: normal navigation and lower-rudder fault navigation. Five specific verification targets are defined, spanning lake-experiment validation, software-in-the-loop simulation validation, and an LLM module ablation study.

\begin{enumerate}
  \item \textbf{Normal navigation performance:} verify that the LASSA framework can plan and execute a closed-loop navigation trajectory under nominal actuator conditions, maintaining cross-track error $e_p$ and heading error $e_\psi$ within acceptable bounds throughout the full outbound-and-return mission.
  \item \textbf{Fault detection and triggering:} verify that the perception planning module correctly identifies the navigation abnormality caused by the lower-rudder fault and raises the confirmed fault flag $\hat{\alpha}(t) = 1$ within a bounded detection latency.
  \item \textbf{Fault-tolerant replanning and execution:} verify that once the fault flag is raised, the LLM generates a physically feasible replanned trajectory, the solver certifies it against the boundary, speed, and turning-radius constraints defined in Section~\ref{sec:solver}, and the UUV successfully completes the remainder of the mission under the updated route.
  \item \textbf{Simulation validation of the closed-loop strategy:} verify that the LASSA framework completes the full perception–replanning–execution cycle in a software-in-the-loop simulation environment under continuous sensor noise across three representative fault scenarios---a steering-locked actuator fault, a cross-current environmental disturbance, and a DVL sensor-degradation fault---thereby decoupling the fault-response logic from the physical hardware and enabling systematic parameter sweeps.
  \item \textbf{LLM module ablation:} characterise the sensitivity of trajectory planning quality to (i) the choice of LLM backbone across four candidate models under a fixed prompt and (ii) the quality of prompt engineering across three prompt variants under a fixed model, with the objective of selecting the backbone and prompt configuration deployed in the lake experiments.
\end{enumerate}

\subsection{Experimental Platform}

\subsubsection{UUV}

The UUV used in the experiments is a small autonomous underwater vehicle whose principal specifications are listed in Table~\ref{tab:uuv_specs}. The vehicle is equipped with a tail-mounted four-fin cross-rudder arrangement (two horizontal and two vertical fins, crossing angle $90^\circ$). In the fault scenario, the lower vertical rudder is physically removed before launch to simulate complete lower-rudder failure.

\begin{table}[!ht]
  \centering
  \caption{Principal specifications of the experimental UUV}
  \label{tab:uuv_specs}
  \begin{tabular}{lll}
    \toprule
    Parameter & Value & Unit \\
    \midrule
    Body length          & 1830  & mm \\
    Body diameter        & 200   & mm \\
    Total weight         & 45.1    & kg \\
    Maximum diving depth & 30    & m  \\
    Maximum speed        & 10    & kn \\
    Rudder configuration & Four-fin cross-rudder & --- \\
    Rudder angle range   & $\pm 60$ & $^\circ$ \\
    Rudder response time & $\leq 0.5$ & s \\
    Rudder servo rated torque & 15 & N$\cdot$m \\
    \bottomrule
  \end{tabular}
\end{table}

The propulsion system uses a tail-mounted three-blade fixed-pitch propeller with a NACA0012 symmetric airfoil profile. Its main parameters are given in Table~\ref{tab:propeller_specs}.

\begin{table}[!ht]
  \centering
  \caption{Propeller and motor specifications}
  \label{tab:propeller_specs}
  \begin{tabular}{lll}
    \toprule
    Parameter & Value & Unit \\
    \midrule
    Blade diameter       & 0.22  & m \\
    Rated speed          & 1200  & r/min \\
    Rated thrust         & 180   & N \\
    Rated torque         & 12    & N$\cdot$m \\
    Propulsion efficiency & $\geq 0.70$ & --- \\
    Motor supply voltage & 24    & V (DC brushless) \\
    Speed control accuracy & $\pm 5$ & r/min \\
    Acceleration response & $\leq 1$ & s \\
    \bottomrule
  \end{tabular}
\end{table}


\subsubsection{Onboard Sensor Suite}

The UUV carries five sensors that collectively provide the measurements $\mathbf{z}_c(t)$ and $\delta_r^m(t)$ defined in \eqref{eq:meas_vec}. Their types and key accuracy parameters are summarized in Table~\ref{tab:sensor_specs}.

\begin{center}
  \includegraphics[width=0.8\linewidth]{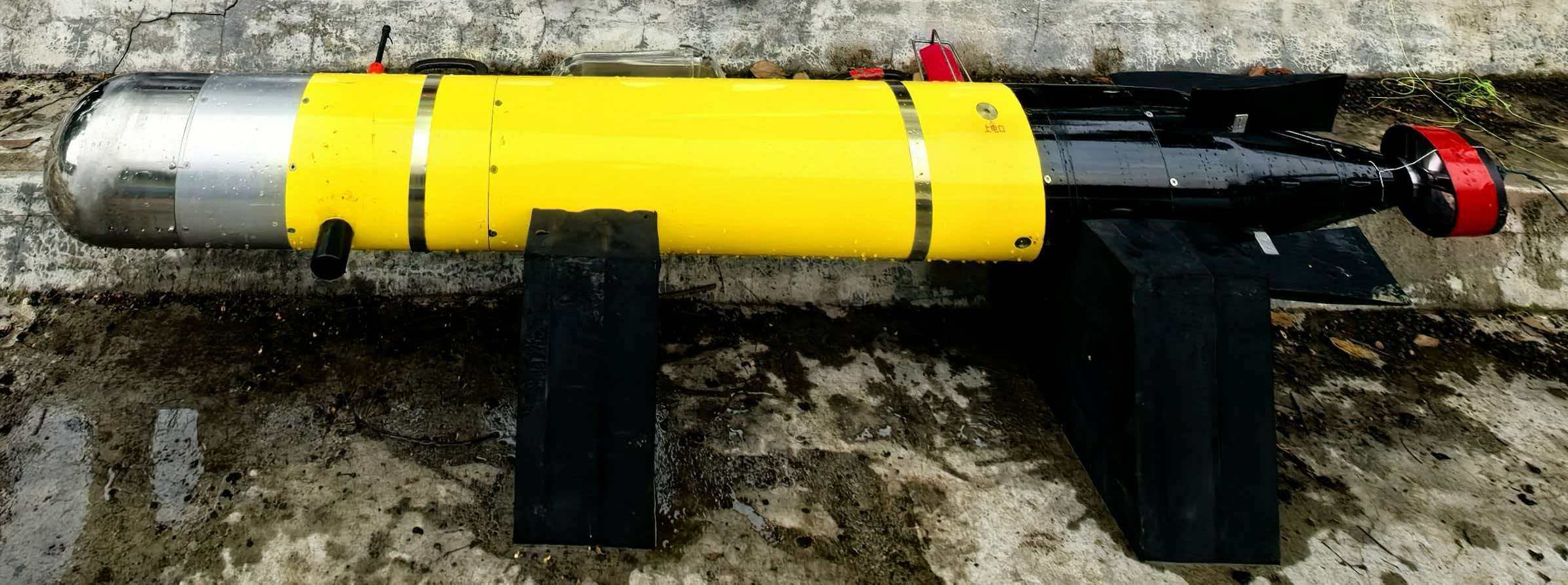}
  \captionof{figure}{Overview of the experimental UUV}
  \label{fig:uuv_basic_info}
\end{center}

\begin{table}[!ht]
  \centering
  \caption{Onboard sensor specifications (all outputs via CAN bus)}
  \label{tab:sensor_specs}
  \begin{tabular}{lll}
    \toprule
    Sensor & Quantity & Accuracy \\
    \midrule
    GPS receiver        & Position $(x^m, y^m)$ & $\pm$10~m \\
    Compass / IMU       & Heading $\psi^m$      & $\pm1^\circ$ \\
    Pressure sensor     & Depth $d^m$           & $\pm$0.01~m \\
    Speed log           & Speed $u^m$           & $\pm$1~kn \\
    Rudder-state sensor & Health $\delta_r^m$   & Boolean \\
    \bottomrule
  \end{tabular}
\end{table}

All sensor signals are encoded as CAN frames and transmitted to the shore-based agent at a unified sampling rate of $f_s = 10$ Hz via a 4G/5G wireless link.

\subsubsection{Intelligent Agent and LLM Deployment}

For experimental convenience, the LASSA agent is deployed on a shore-based computing platform that communicates with the UUV in real time over a wireless link. The platform runs the agent perception–planning–scheduling cycle and hosts the LLM inference engine. An open-source LLM with a parameter scale in the range of 10B–30B is used. The single-instruction inference latency is $\leq$500 ms, satisfying the real-time requirement of the fault-response pipeline. The shore-based monitoring interface displays the live GPS track, the reference trajectory, and the agent status in real time; representative screenshots are presented and discussed in Section~5. The shore-based deployment is adopted here only for experimental convenience and monitoring; the LASSA architecture itself is designed to be portable to onboard computing without modification, and onboard deployment is targeted as part of future work.

\subsubsection{Simulation Platform}
\label{sec:sim_platform}

To enable systematic closed-loop validation of the LASSA control logic prior to lake deployment, and to decouple the fault-response pipeline from the physical hardware, a software-in-the-loop simulation platform was constructed. The platform comprises five modules that communicate via TCP over a local loopback connection, with the simulated UUV acting as server and the agent acting as client.

The \textit{simulated UUV module} executes a discrete-time kinematic model at a fixed state-report period of $T_s = 2.0$~s and a nominal surge speed of $v = 2.0$~m/s. At each cycle it advances the vehicle position and heading, applies sensor noise described below, and transmits the noisy state vector to the agent. The \textit{agent module} performs local coordinate frame initialisation, cross-track deviation monitoring, LLM context assembly, solver invocation, and command dispatch, mirroring the same perception–planning–scheduling logic used in the lake experiments. The \textit{LLM planning module} generates trajectory waypoints through a hybrid approach combining geometric arc construction with LLM-based waypoint interpolation; the LLM backbone is the model selected in the ablation study (Section~\ref{sec:ablation_design}). The \textit{solver module} checks each candidate path against boundary constraints and computes cross-track deviation from the active reference polyline. The \textit{visualisation module} renders the live position track, reference path, depth profile, replanning results, and per-phase latency on a shared display.

To replicate the continuous measurement uncertainties present in the lake experiments, the simulated state report incorporates two noise models. Lateral position error follows a random-walk-plus-white-noise model with a maximum drift amplitude of 3~m, a walk strength of 0.25~m per step, and superimposed white noise $\sigma_\mathrm{lat} = 0.08$~m. Depth observation error is strictly positive (biased upward), modelled with a maximum offset of 0.35~m, a walk strength of 0.04~m per step, and superimposed white noise $\sigma_d = 0.02$~m. These models are qualitatively consistent with the GPS and pressure-sensor accuracy figures listed in Table~\ref{tab:sensor_specs}.

\subsection{Test Environment}

The experiments were conducted in a calm inland lake. The key environmental parameters are summarized in Table~\ref{tab:env_params}. The test area is a closed freshwater lake free of tidal currents, and wireless communication (4G/5G/WiFi) covered the entire test area without blind spots. Within the UUV operating depth range of 0–0.3 m, the water depth was spatially uniform, so no depth-collision avoidance was required during navigation.

\begin{table}[!ht]
  \centering
  \caption{Test environment parameters}
  \label{tab:env_params}
  \begin{tabular}{ll}
    \toprule
    Parameter & Value \\
    \midrule
    Water body type          & Inland closed lake \\
    Avg.\ / max.\ / min.\ depth & 1.3 / 2.0 / 1.0~m \\
    UUV operating depth      & 0--0.3~m \\
    Current / wave condition & Still water / none \\
    Communication coverage   & Full (4G/5G/WiFi) \\
    \bottomrule
  \end{tabular}
\end{table}


\begin{center}
  \includegraphics[width=0.8\linewidth]{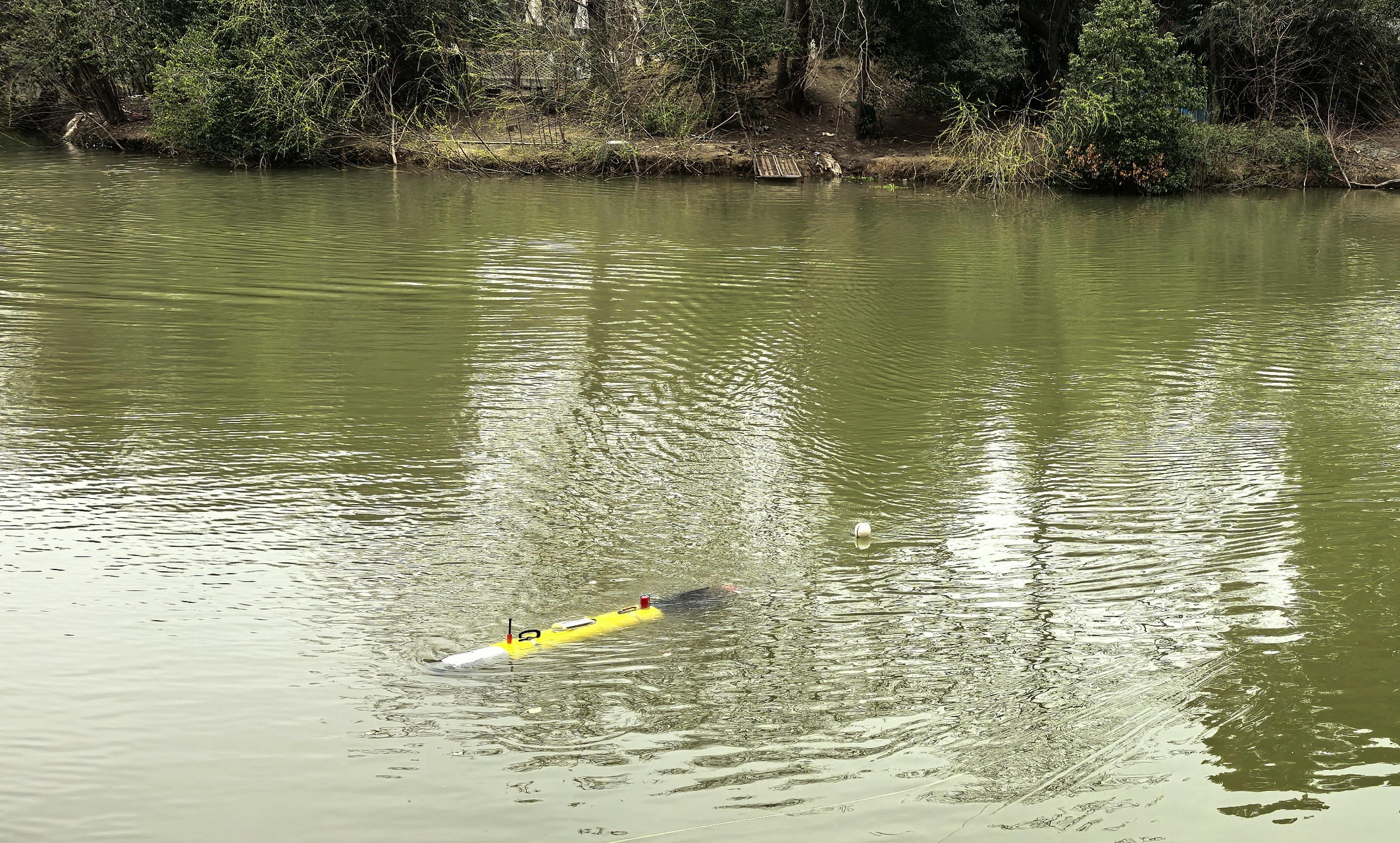}
  \captionof{figure}{UUV navigation status in the test lake}
  \label{fig:uuv_lake_state}
\end{center}

\subsection{Experimental Configurations}

Two experimental configurations are defined. The \textbf{normal navigation experiment} (Exp-N) serves as the performance baseline under nominal actuator conditions. The \textbf{lower-rudder fault experiment} (Exp-F) introduces the rudder fault before launch and tests the full fault-detection, replanning, and execution pipeline. Table~\ref{tab:exp_configs} summarises the key parameters of both configurations.

\begin{table}[!ht]
  \centering
  \caption{Experimental configurations and key parameters}
  \label{tab:exp_configs}
  \begin{tabular}{lll}
    \toprule
    Parameter & Exp-N & Exp-F \\
    \midrule
    Rudder condition       & Nominal  & Lower rudder removed \\
    Turning radius (init.) & 7~m      & 4~m \\
    Turning radius (repl.) & ---      & 12~m \\
    Navigation speed (init.)& 2~kn   & 2~kn \\
    Navigation speed (repl.)& ---    & 1~kn \\
    Forward distance       & 80~m    & 90~m \\
    Threshold $\varepsilon_p$ & 3.0~m & 3.0~m \\
    Window $n_w$           & 5 cycles & 5 cycles \\
    Accept.\ radius $r_\mathrm{acc}$ & 2~m & 2~m \\
    \bottomrule
  \end{tabular}
\end{table}

In Exp-F, the initial turning radius of 4 m is deliberately set below the fault-state minimum turning radius $R_\mathrm{min}^\mathrm{fault}$ (estimated at approximately 10 m) to ensure that the navigation abnormality is triggered reliably. The replanned turning radius of 12 m satisfies constraint \eqref{eq:rmin_combined} and was verified by the solver before deployment.

The simulation experiment (Exp-S) comprises three sub-scenarios: a steering-lock actuator fault, a cross-current environmental disturbance, and a DVL sensor-degradation fault. The cross-current and DVL sub-scenarios share the same SIL platform and noise models but use sub-scenario-specific fault and parameter settings, which are introduced together with their results in Section~\ref{sec:exps}; only the steering-lock sub-scenario uses the parameter set summarised in Table~\ref{tab:sim_configs}. In the steering-lock sub-scenario, the fault is modelled through a \textit{steering-lock mechanism}: upon receipt of a replanned trajectory, the simulated UUV does not immediately alter its heading. Instead, it maintains its current heading and dives at a rate limited to $\Delta d_\mathrm{max} = 0.2$~m per cycle until the accumulated depth increment reaches the unlock threshold $\Delta d_\mathrm{unlock} = 0.2$~m, after which waypoint tracking is enabled. This mechanism approximates the physical scenario in which the lower vertical rudder is inoperative and the vehicle must submerge until the upper control surface re-enters effective flow before yaw authority is restored. The cross-current and DVL sub-scenarios do not impose this dive-then-unlock dependency.

\begin{table}[!ht]
  \centering
  \caption{Key parameters of the steering-lock sub-scenario in Exp-S}
  \label{tab:sim_configs}
  \begin{tabular}{ll}
    \toprule
    Parameter & Value \\
    \midrule
    Coord.\ range (lat./fwd./depth) & $[\pm100]$/$[\pm100]$/$[\pm2]$~m \\
    Mission target (local)          & $(0,\ 90,\ 0)$~m \\
    Recovery point (local)          & $(0,\ 0,\ 0)$~m \\
    Nominal speed                   & 2.0~m/s \\
    State report period $T_s$       & 2.0~s \\
    Cross-track threshold           & 10~m \\
    Initial turning radius          & 5~m \\
    Replanned turning radius        & 10~m \\
    Waypoint acceptance radius      & 10~m \\
    Steering-unlock delta $\Delta d_\mathrm{unlock}$ & 0.2~m \\
    Max.\ depth step $\Delta d_\mathrm{max}$         & 0.2~m \\
    Replan dive depth               & 0.5~m \\
    Replan forward offset           & 5.0~m \\
    Initial path speed              & 2.0~m/s \\
    Replanned path speed            & 1.0~m/s \\
    Lateral drift amplitude         & 3.0~m \\
    Lateral drift walk strength     & 0.25~m/step \\
    Depth bias max offset           & 0.35~m \\
    \bottomrule
  \end{tabular}
\end{table}

\subsection{Experimental Procedure}

The experimental procedure is identical for both configurations and consists of four stages.

\textbf{Stage 1 – Pre-launch preparation.} The UUV hardware is inspected (sensing system, propulsion system, and rudder mechanism). For Exp-F, the lower rudder is physically removed at this stage. The LLM and agent software are initialized on the shore-based platform; mission parameters from Table~\ref{tab:exp_configs} are loaded; and the wireless communication link is verified at both ends.

\textbf{Stage 2 – Launch and initial navigation.} The UUV is placed in the water at the designated launch point. The LASSA control loop is started, and the LLM generates the initial trajectory (turning radius and waypoint sequence per Table~\ref{tab:exp_configs}). The agent begins closed-loop navigation control. The shore-based monitor logs position, heading, speed, and depth at $f_s = 10$ Hz throughout the experiment.

\textbf{Stage 3 – Fault detection and replanning (Exp-F only).} Once the agent perception planning module raises the confirmed fault flag $\hat{\alpha}(t) = 1$, the fault context package $\mathcal{I}(t)$ is assembled and passed to the LLM. The LLM generates a replanned strategy $\Theta(t)$, which is checked by the solver against the three constraints in Section~\ref{sec:solver}. After the solver issues a \textsc{pass} verdict, the new trajectory commands are forwarded to the UUV actuators.

\textbf{Stage 4 – Return and recovery.} The UUV completes the outbound leg and executes the return route to the designated recovery point. After surfacing, it is recovered and the logged data are downloaded for analysis.

\subsection{LLM Ablation Study Design}
\label{sec:ablation_design}

To select the LLM backbone and prompt template used in both the lake and simulation experiments, and to provide a systematic characterisation of how each factor affects trajectory planning quality, an ablation study was conducted across two orthogonal groups before the main experiments.

\textbf{Group~1 — Model comparison under a fixed prompt.} Four LLM backbones were evaluated: Kimi~K2.5, Qwen3-max, DeepSeek-V3.2, and GPT-5.3. Each model received an identical structured prompt specifying the navigable area boundary, start and end waypoints, required arc radius, tangency conditions, and sampling density. Two scenarios were tested. In the \textit{normal scenario}, all geometric constraints are mutually consistent and a valid closed path exists within the boundary. In the \textit{abnormal scenario}, the specified arc radius deliberately exceeds the geometric feasibility limit imposed by the navigable area, creating an intentional constraint conflict. The normal scenario assesses whether a model can generate smooth, boundary-compliant arcs with correct tangency and uniform sampling; the abnormal scenario tests whether the model can identify constraint infeasibility and produce either an explicit infeasibility signal or a best-effort approximation, rather than silently generating a geometrically invalid path.

\textbf{Group~2 — Prompt variant comparison under a fixed model.} Using the Kimi~K2.5 backbone identified from Group~1, three prompt variants of increasing descriptive precision were evaluated. The \textit{minimal variant} specifies only the core geometric parameters (boundary, arc radius, waypoints). The \textit{standard variant} adds structured constraint sections and an explicit output format requirement (comma-separated coordinate tuples, no natural-language explanation). The \textit{detailed variant} further provides step-by-step geometric reasoning guidance, explicit statements of tangency conditions, and per-segment sampling allocation. Each variant was tested under the same normal and abnormal scenarios as Group~1. Path quality was assessed qualitatively by examining arc tangency accuracy, sampling uniformity, boundary clearance, and path closure in the normal scenario, and by whether the model correctly identified or approximated the infeasible constraint in the abnormal scenario.

The model and prompt configuration that achieved the highest geometric fidelity in the normal scenario and the most appropriate constraint-conflict response in the abnormal scenario were adopted as the LLM deployment configuration for Exp-N, Exp-F, and Exp-S.

\section{Results and Analysis}

\subsection{Normal Navigation Experiment (Exp-N)}

\subsubsection{Navigation Performance}

Under the nominal actuator conditions of Exp-N, the UUV completed the full outbound-and-return mission without triggering any fault flag. The initial trajectory generated by the LLM specified a turning radius of $R = 7$~m, which falls within the target planning range of $6 \pm 1$~m specified in the mission parameters, a navigation speed of $u = 2$~kn, and a forward distance of 80~m. The agent perception planning module evaluated the estimated state at each 10~Hz cycle and produced $\hat{\alpha}(t) = 0$ throughout the entire run, confirming that no navigation abnormality occurred.

\subsubsection{Quantitative Results}

The key navigation metrics recorded during Exp-N are summarized in Table~\ref{tab:normal_results}. The fault flag $\hat{\alpha}$ remained at zero throughout the run and the mission was completed in full, confirming that the LASSA framework operates correctly under nominal conditions. The measured maximum cross-track error $e_p^\mathrm{max} = 2.8$~m remains below the fault-detection threshold $\varepsilon_p = 3.0$~m. These results provide a reliable performance baseline for the fault experiment.

\begin{table}[!ht]
  \centering
  \caption{Key navigation metrics for Exp-N (normal navigation)}
  \label{tab:normal_results}
  \begin{tabular}{lll}
    \toprule
    Metric & Value & Requirement \\
    \midrule
    LLM-planned turning radius   & 7~m   & $6 \pm 1$~m \\
    Navigation speed             & 2~kn  & 2~kn \\
    Forward distance             & 80~m  & 80~m \\
    Max.\ cross-track error $e_p^\mathrm{max}$ & 2.8~m  & $< 3.0$~m \\
    Fault flag $\hat{\alpha}$ raised & No  & No \\
    Mission completed            & Yes   & Yes \\
    \bottomrule
  \end{tabular}
\end{table}

\begin{center}
  \includegraphics[width=0.8\linewidth]{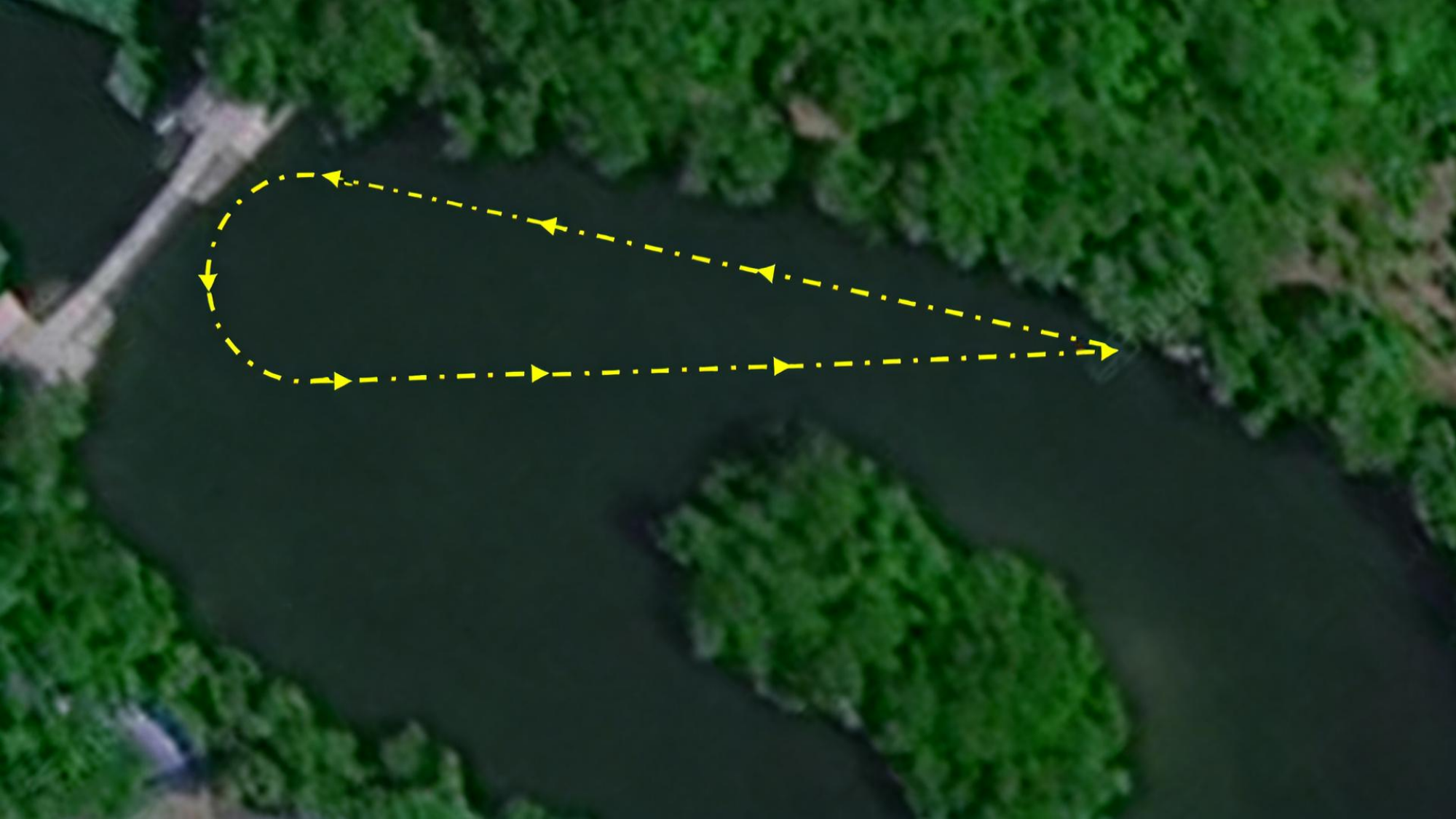}
  \captionof{figure}{Aerial view of the complete Exp-N mission track, showing the outbound leg (right-to-left) and the return leg along the planned route with no deviation throughout the run}
  \label{fig:normal_navigation_aerial}
\end{center}

\subsection{Lower-Rudder Fault Experiment (Exp-F)}

\subsubsection{Initial Navigation and Abnormality Detection}

In Exp-F, the lower rudder was removed before launch to simulate complete lower-rudder failure. The LLM generated an initial route with a turning radius of $R_0 = 4$~m and a speed of $u_0 = 2$~kn, which is deliberately below the estimated fault-state minimum turning radius $R_\mathrm{min}^\mathrm{fault} \approx 10$~m. As a result, once the UUV reached the first turning waypoint, it was unable to follow the planned arc, and the cross-track error $e_p(t)$ exceeded the threshold $\varepsilon_p = 3.0$~m for $n_w = 5$ consecutive sampling cycles. The perception planning module therefore raised the confirmed fault flag $\hat{\alpha}(t) = 1$, triggering the LLM reasoning pipeline.

\subsubsection{LLM-Driven Trajectory Replanning}

Upon receiving the context package $\mathcal{I}(t)$, the LLM performed chain-of-thought reasoning and proposed a replanned strategy with a turning radius of $R_\mathrm{new} = 12$~m and a reduced speed of $u_\mathrm{new} = 1$~kn. This strategy was submitted to the physical solver, which executed the three sequential checks described in Section~\ref{sec:solver}. The solver verified that:
\begin{enumerate}
  \item \textbf{Boundary check:} all arc sample points of the replanned path maintained a clearance $d_b \geq d_\mathrm{safe}$ from the lake boundary;
  \item \textbf{Speed constraint:} $u_\mathrm{new} = 1$~kn satisfies $u_\mathrm{min} \leq u_\mathrm{new} \leq u_\mathrm{max}$;
  \item \textbf{Turning-radius constraint:} $R_\mathrm{new} = 12~\text{m} \geq R_\mathrm{min}^\mathrm{fault} \approx 10~\text{m}$ per \eqref{eq:rmin_combined}.
\end{enumerate}
All three checks passed on the first solver invocation, and the solver returned a \textsc{pass} verdict. The verified command list $\mathbf{C}(t)$ was forwarded to the UUV via the scheduling module.

\subsubsection{Post-Replanning Navigation and Return}

After receiving the updated commands, the UUV resumed navigation according to the replanned trajectory. With the larger turning radius and reduced speed, the vehicle was able to execute the required turns despite the degraded yaw authority caused by the missing lower rudder. The UUV completed the remaining outbound leg and successfully executed the return operation to the designated recovery point.

The visible deviation in the initial turning segment, followed by the recovery to the replanned route, visually confirms the fault-detection and replanning behavior; the aerial view of the complete mission track is shown in Figure~\ref{fig:fault_navigation_state_3}.

\subsubsection{Quantitative Results and Comparative Summary}

The key parameters of Exp-F are summarized in Table~\ref{tab:fault_results}, and a side-by-side comparison of the two experiments is given in Table~\ref{tab:comparison}.

\begin{table}[!ht]
  \centering
  \caption{Key metrics for Exp-F (lower-rudder fault experiment)}
  \label{tab:fault_results}
  \begin{tabular}{ll}
    \toprule
    Metric & Value \\
    \midrule
    Initial turning radius $R_0$              & 4~m \\
    Initial navigation speed $u_0$            & 2~kn \\
    Forward distance                          & 90~m \\
    Fault-state min.\ radius $R_\mathrm{min}^\mathrm{fault}$ & $\approx$10~m \\
    Replanned turning radius $R_\mathrm{new}$ & 12~m \\
    Replanned speed $u_\mathrm{new}$          & 1~kn \\
    Solver checks passed (1st attempt)        & 3/3 \\
    Fault flag raised                         & Yes \\
    Mission completed after replanning        & Yes \\
    \bottomrule
  \end{tabular}
\end{table}

\begin{table}[!ht]
  \centering
  \caption{Comparative summary of Exp-N and Exp-F}
  \label{tab:comparison}
  \begin{tabular}{lll}
    \toprule
    Item & Exp-N & Exp-F \\
    \midrule
    Rudder condition            & Nominal  & Lower rudder removed \\
    Turning radius (init.)      & 7~m      & 4~m \\
    Turning radius (exec.)      & 7~m      & 12~m (replanned) \\
    Navigation speed (init.)    & 2~kn     & 2~kn \\
    Navigation speed (exec.)    & 2~kn     & 1~kn (replanned) \\
    Fault flag raised           & No       & Yes \\
    LLM replanning triggered    & No       & Yes \\
    Solver invocations          & 0        & 1 \\
    Solver pass on 1st attempt  & ---      & Yes \\
    Mission completed           & Yes      & Yes \\
    \bottomrule
  \end{tabular}
\end{table}

The experimental results demonstrate that the proposed LASSA framework successfully achieves the three lake-experiment verification targets stated in Section~4.1. Under normal conditions, the UUV tracks the planned trajectory throughout the mission with no false fault alarms. Under the lower-rudder fault condition, the framework correctly detects the navigation abnormality, invokes the LLM to generate a physically feasible replanned route, obtains first-attempt solver certification, and guides the vehicle to complete the mission. The coordinated operation of the perception planning module, the LLM reasoning engine, the physical solver, and the actuator control layer confirms the practical engineering feasibility of the LASSA architecture for fault-state UUV autonomous navigation.

\begin{center}
  \includegraphics[width=0.8\linewidth]{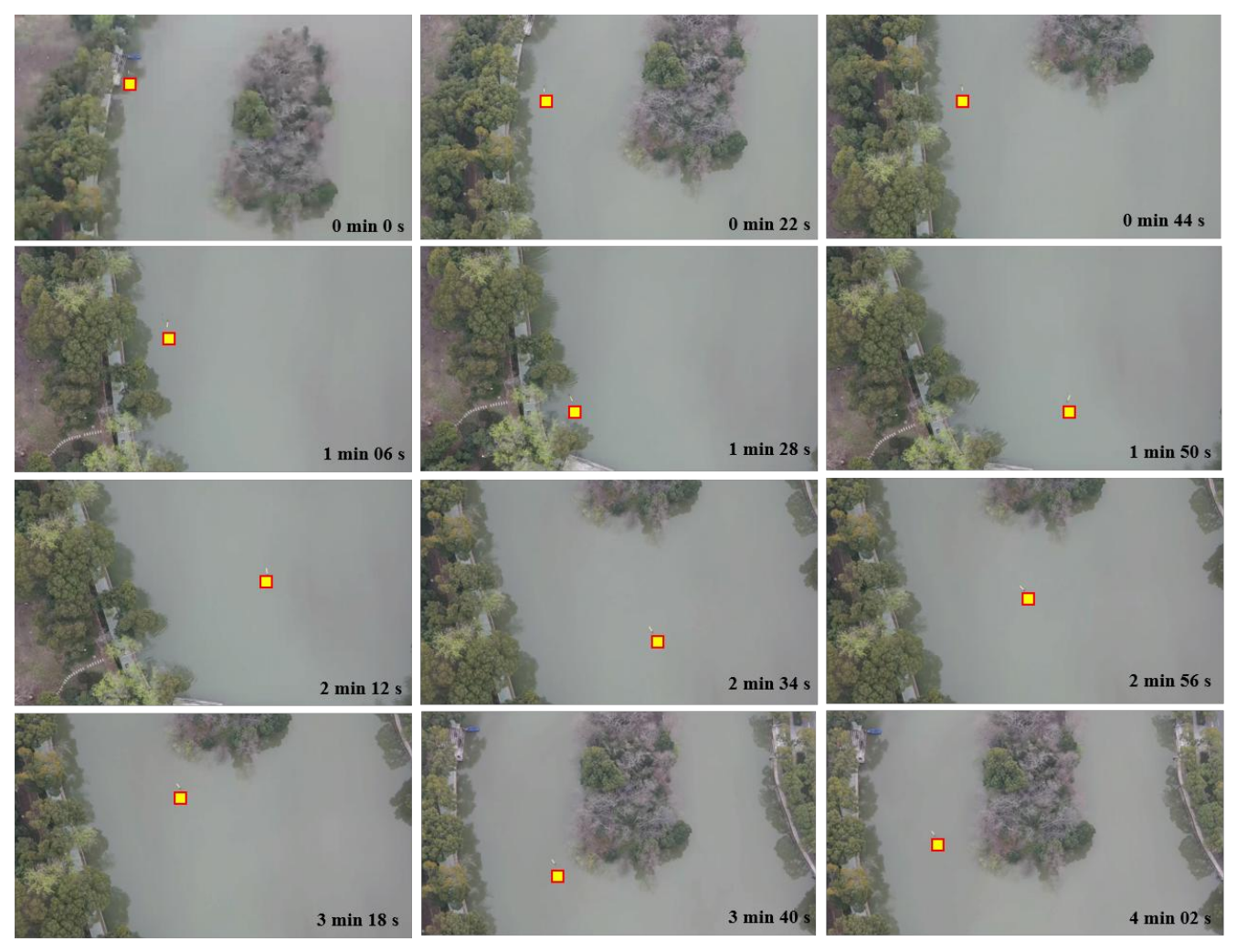}
  \captionof{figure}{Aerial view of the complete Exp-F mission track, showing the initial deviation, fault-triggered replanning point, and the subsequent return route}
  \label{fig:fault_navigation_state_3}
\end{center}

\subsection{Simulation Experiment (Exp-S)}
\label{sec:exps}
 
\subsubsection{Initialisation and Straight Navigation}
 
\begin{figure}[!ht]
  \centering
  \includegraphics[width=\linewidth]{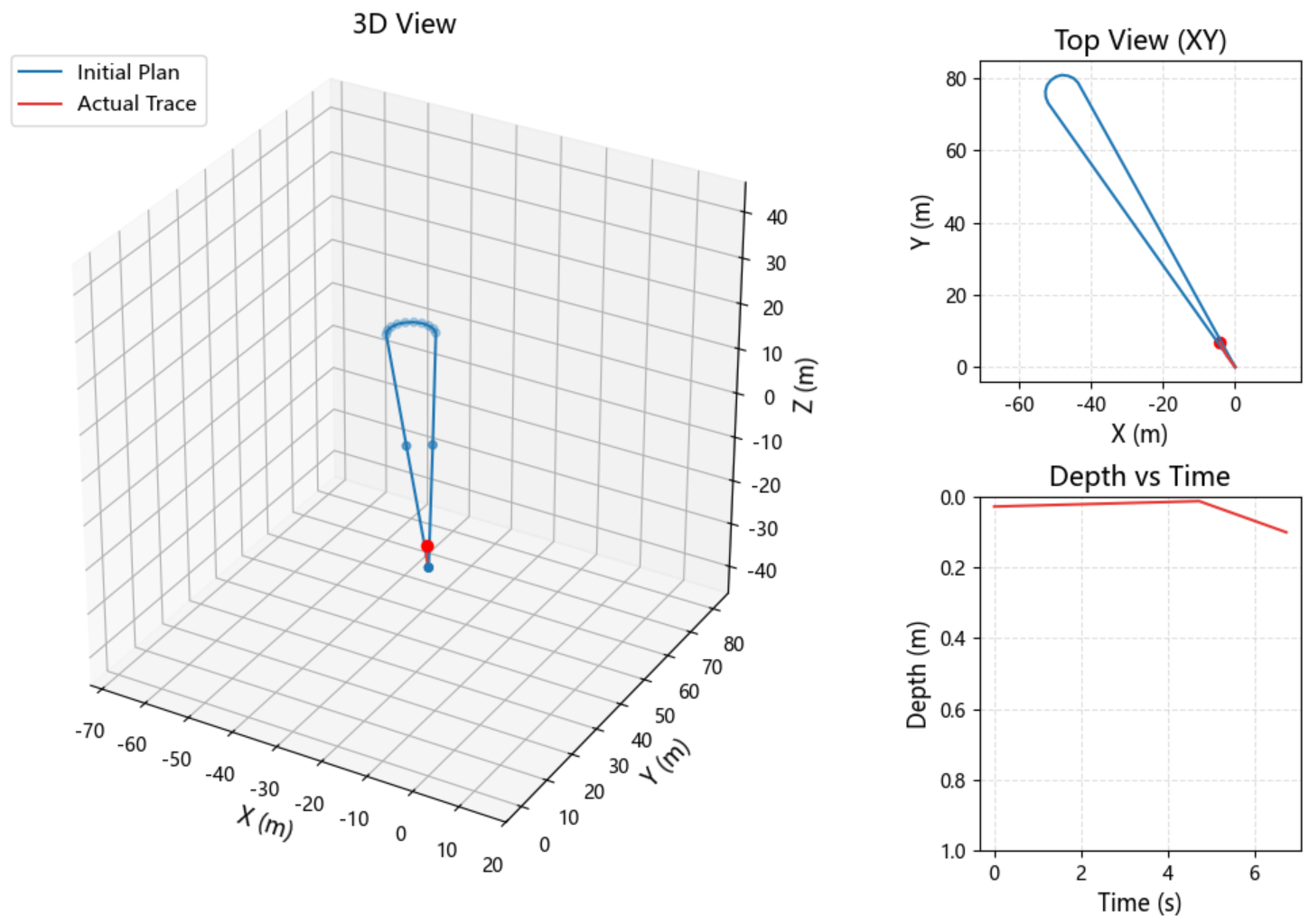}\\[4pt]
  \includegraphics[width=\linewidth]{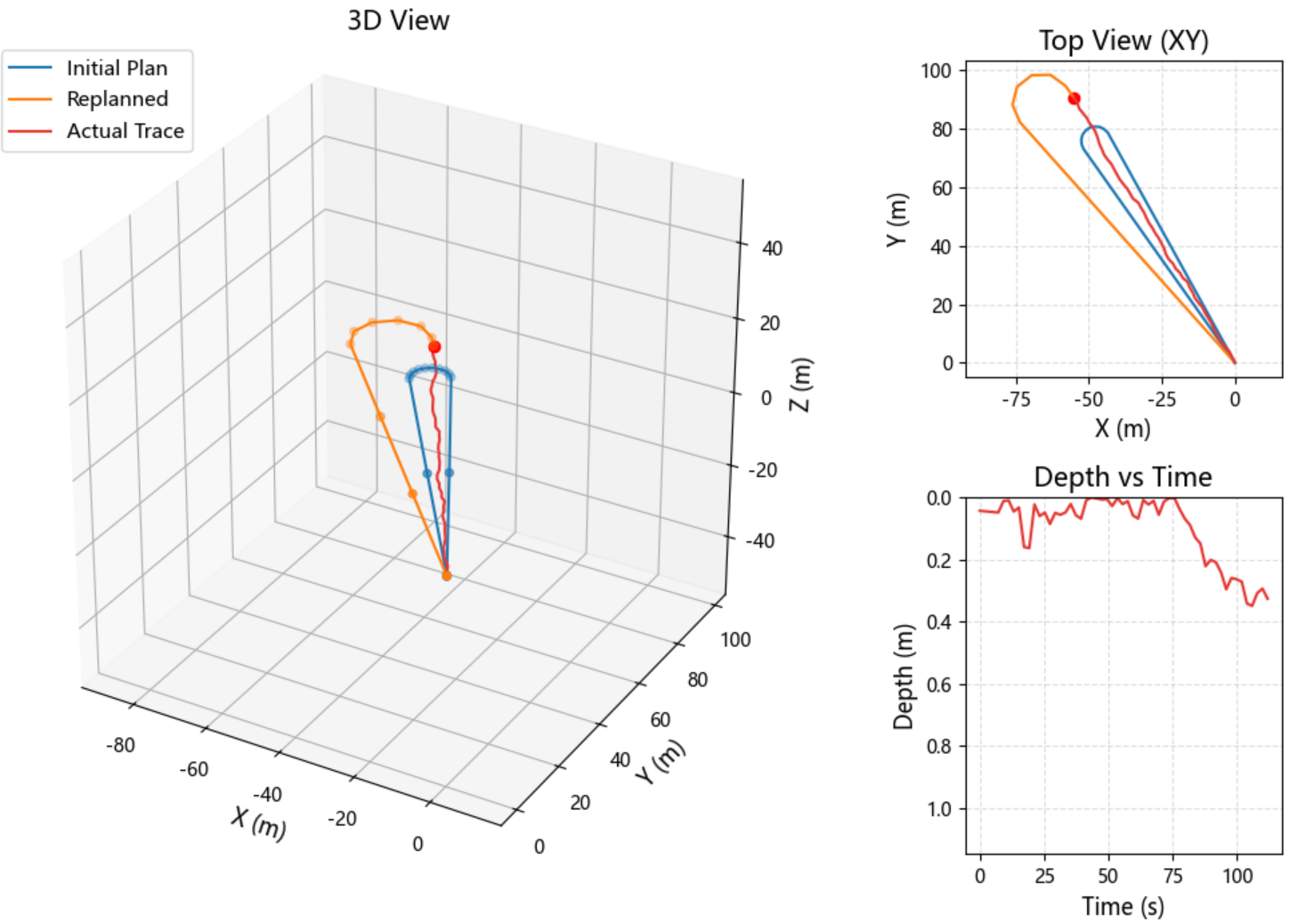}\\[4pt]
  \includegraphics[width=\linewidth]{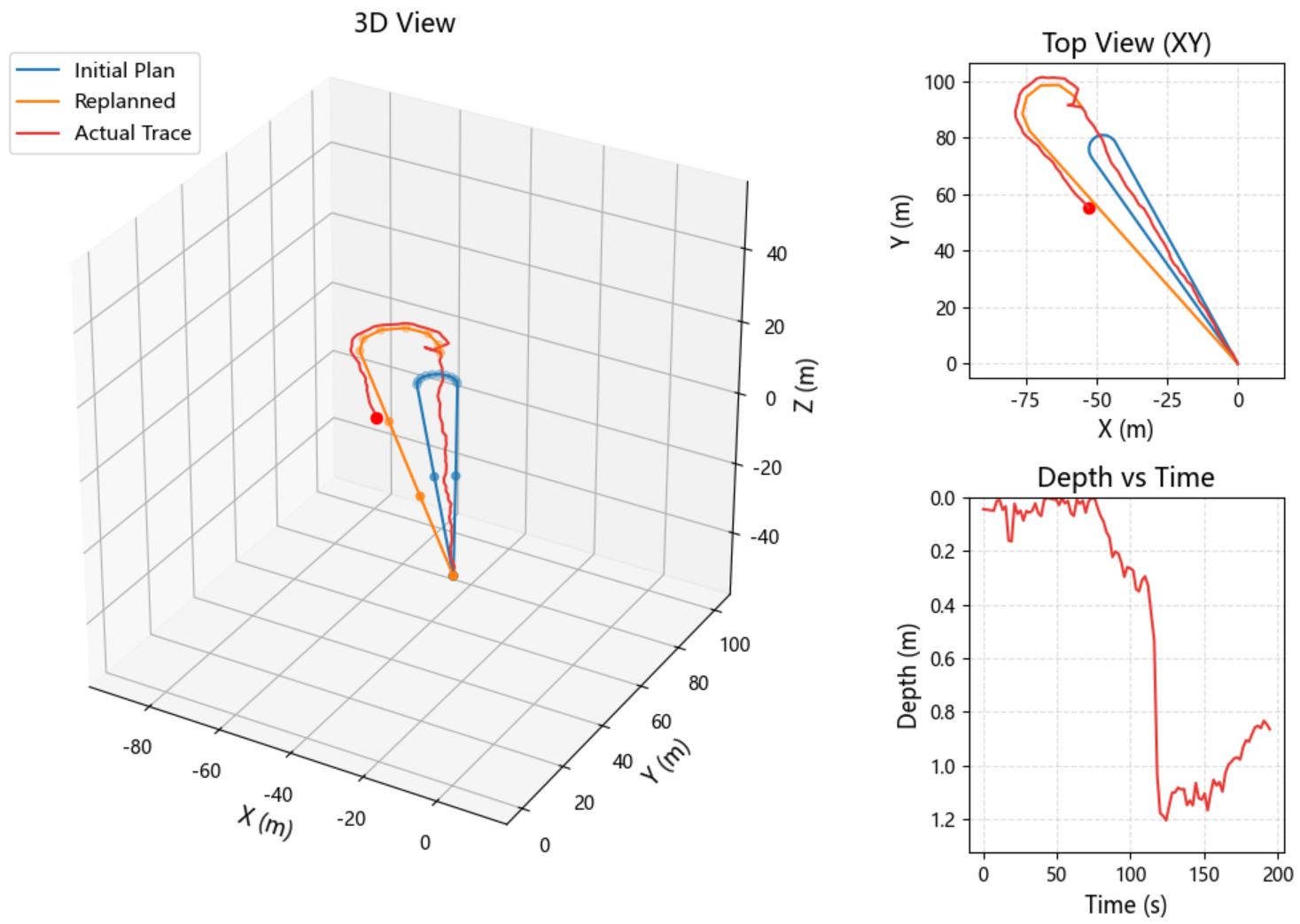}
  \caption{Exp-S three-phase trajectory results. \textit{Top}: Phase~1, path initialisation and straight navigation (stable depth, initial route blue, actual trace red). \textit{Middle}: Phase~2, deviation detection and LLM-driven replanning (blue: initial path; orange: replanned route; red: actual trace; depth begins to fluctuate). \textit{Bottom}: Phase~3, steering-lock dive and return execution (depth reaches $\approx$1.2~m during the lock phase before recovering).}
  \label{fig:sim_phases}
\end{figure}

At mission start the agent initialised the local coordinate frame with the UUV launch point as origin and instructed the LLM to generate an initial path to the mission target at $(0, 90, 0)$~m with a turning radius of 5~m. Rather than tracking the arc sequence point-by-point, the simulated UUV extracted the outbound heading from the initial plan and advanced at 2.0~m/s along that heading in a straight line. This design intentionally creates a scenario in which the reported position (subject to continuous lateral drift up to 3~m) progressively diverges from the planned polyline as the vehicle advances, establishing the precondition for deviation detection without requiring a pre-specified fault trigger time. The 3D trajectory view, top-view projection, and depth profile for the three phases of Exp-S are shown in Figure~\ref{fig:sim_phases}.
 
\subsubsection{Cross-Track Deviation Detection and Replanning}
 
As the simulated UUV continued along its initial heading, the solver computed the perpendicular distance from the noisy reported position to the reference polyline at each $T_s = 2.0$~s cycle. Once this cross-track deviation exceeded the threshold of 10~m, the agent assembled the context package from the current state estimate and passed it to the LLM. The LLM generated a return path to the recovery point $(0, 0, 0)$~m with a replanned turning radius of 10~m and a reduced waypoint speed of 1.0~m/s. The solver verified the replanned path against the boundary constraints and confirmed feasibility before the agent dispatched the waypoint sequence to the simulated UUV. As shown in the middle panel of Figure~\ref{fig:sim_phases}, the orange replanned route and the diverged red actual trace are both visible, confirming that the replanning was triggered at the correct deviation point.
 
\subsubsection{Steering-Lock Dive and Return Execution}
 
Upon receipt of the replanned trajectory, the simulated UUV entered the steering-lock phase, maintaining its current heading while diving at a rate capped at $\Delta d_\mathrm{max} = 0.2$~m per cycle. Once the accumulated depth increment reached $\Delta d_\mathrm{unlock} = 0.2$~m, the steering lock was released and the vehicle transitioned to waypoint-tracking mode. The depth profile in the bottom panel of Figure~\ref{fig:sim_phases} clearly shows the characteristic dive-then-recover signature: depth increases sharply during the steering-lock phase (reaching approximately 1.2~m) before recovering as the vehicle begins tracking the replanned route. Despite the continuous lateral drift noise present throughout the return leg, the vehicle reached the recovery point and completed the mission.
 
\begin{figure}[!ht]
  \centering
  \includegraphics[width=\linewidth]{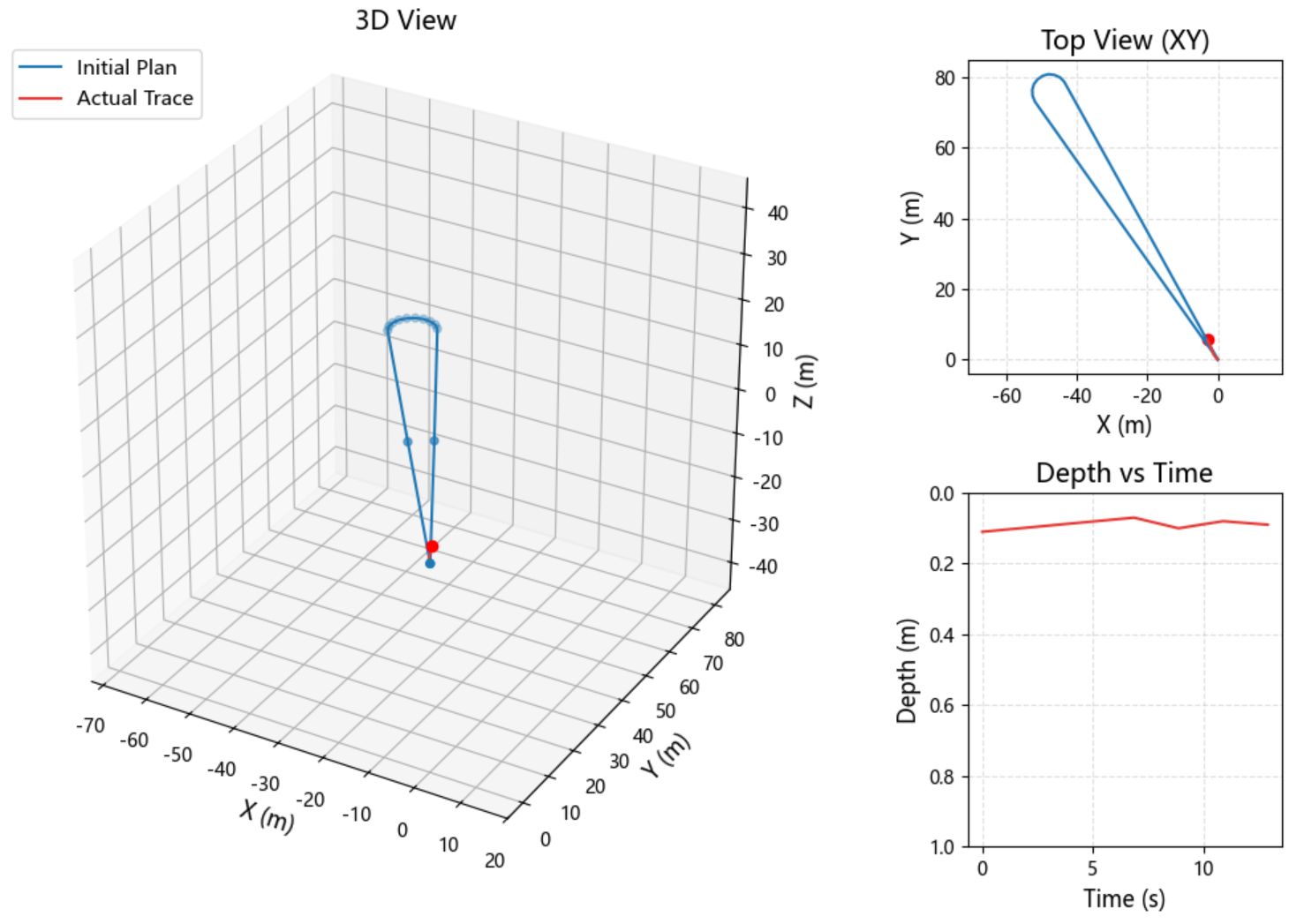}\\[4pt]
  \includegraphics[width=\linewidth]{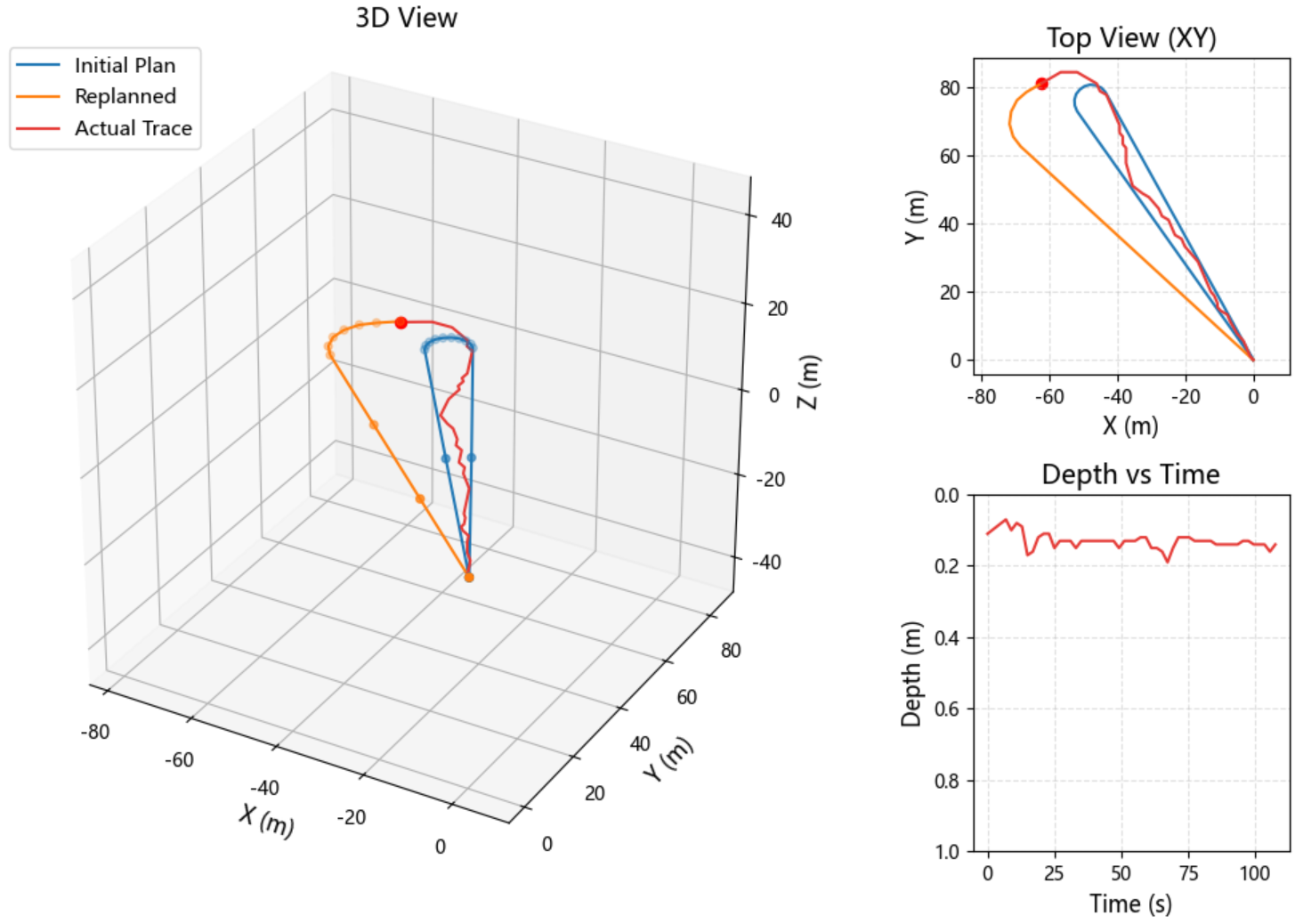}\\[4pt]
  \includegraphics[width=\linewidth]{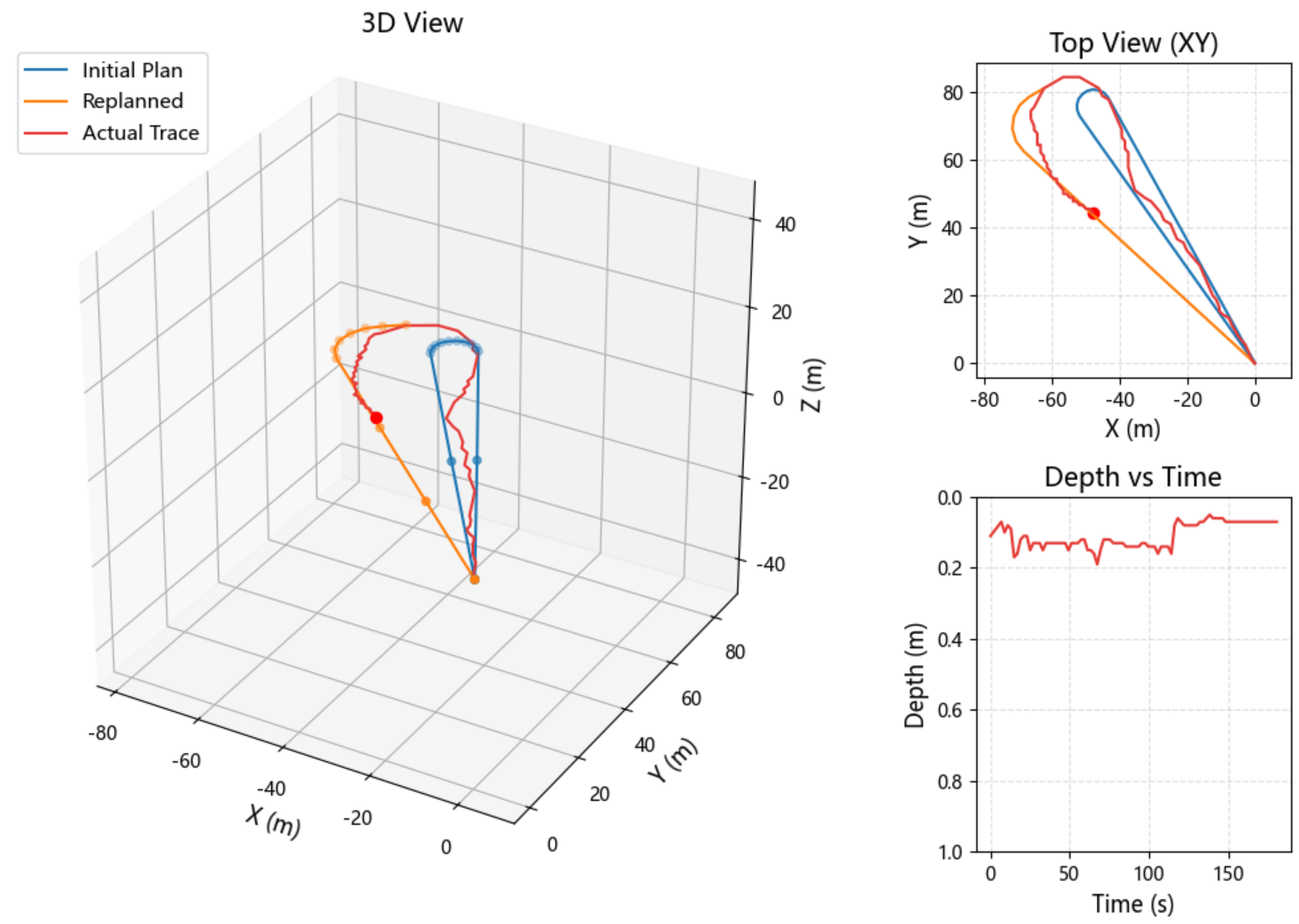}
  \caption{Surface-constrained simulation variant (UUV unable to actively adjust depth). \textit{Top}: Phase~1, path initialisation; depth profile is flat throughout. \textit{Middle}: Phase~2, deviation detection and replanning; depth remains stable (0.1--0.25~m) with no steering-lock dive. \textit{Bottom}: Phase~3, return execution along the replanned path with stable depth, demonstrating that the LASSA pipeline succeeds without dive-assisted recovery.}
  \label{fig:sim_real_phases}
\end{figure}

To further assess the robustness of the LASSA replanning pipeline under a more constrained mobility model, a supplementary simulation was conducted in which the UUV is unable to actively adjust its depth (approximating the surface-constrained operating mode of the lake experiments). Under this constraint, the steering-lock dive phase is suppressed and the vehicle must execute the replanned turn relying solely on the adjusted turning radius without depth-assisted recovery of yaw authority. The trajectory results for this surface-constrained variant are shown in Figure~\ref{fig:sim_real_phases}. The replanning pipeline succeeded in generating a valid return path and guiding the vehicle to the recovery point in both cases, confirming that the LASSA framework is effective across both depth-capable and surface-constrained operating modes.
 
\subsubsection{Robustness Under Cross-Current Disturbance}
 
\begin{figure}[!ht]
  \centering
  \includegraphics[width=\linewidth]{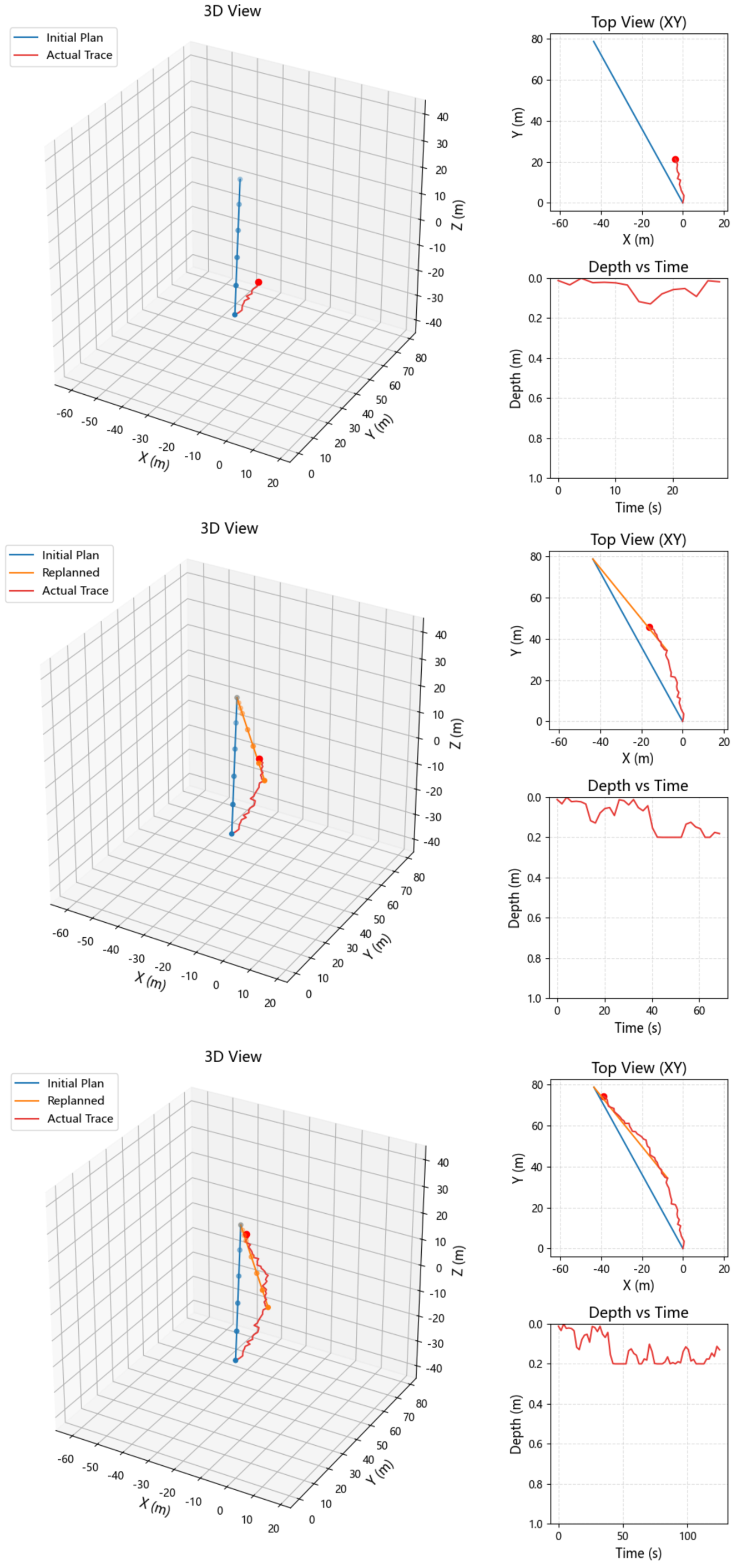}
  \caption{Robustness validation under cross-current disturbance. \textit{Top}: the UUV is progressively displaced laterally by a steady cross-current, with the actual trace diverging from the initial plan. \textit{Middle}: the cross-track deviation triggers LLM replanning; the LLM generates a corrected straight-line path to the mission target (orange) and additionally outputs a rudder-bias correction of $\approx -11.5^\circ$ that accounts for the persistent lateral force. \textit{Bottom}: the UUV executes the replanned path with the applied rudder correction and successfully reaches the mission target against the cross-current.}
  \label{fig:env_phases}
\end{figure}

Beyond the actuator-fault scenarios already presented, the LASSA framework was evaluated under a continuous lateral hydrodynamic disturbance to assess its robustness against environmental perturbations. A simulated cross-current was injected, exerting a steady lateral velocity of $0.3$~m/s perpendicular to the nominal heading throughout the run. The three stages of this experiment are shown in Figure~\ref{fig:env_phases}.
 
In the first stage (top panel), the UUV is progressively pushed off the planned trajectory by the cross-current; the actual trace gradually diverges from the initial plan as the lateral perturbation accumulates. Once the cross-track deviation crosses the detection threshold ($10$~m, reached after approximately $34$~s of accumulated drift), the agent triggers the LLM reasoning pipeline (middle panel); the LLM produces a replanned straight-line path to the mission target and additionally generates a rudder-bias correction of $\approx -11.5^\circ$ that compensates for the persistent lateral force. In the final stage (bottom panel), the UUV follows the replanned path with the applied rudder correction and successfully reaches the mission target against the cross-current.
 
This experiment demonstrates that the LASSA pipeline can extend beyond actuator faults to handle continuous environmental disturbances. Notably, the LLM autonomously inferred both the need for a rudder bias term and its quantitative value ($\approx -11.5^\circ$) from the deviation signature alone---a corrective action that would require explicit hand-coded logic in conventional rule-based fault-handling systems. The rudder-bias term lies outside the geometric strategy schema $\Theta(t)$ defined in Section~\ref{sec:llm}, and is admitted only after a parameter-bound check confirms it falls within the safe deflection range of the heading controller. The detailed metrics are listed in Table~\ref{tab:env_results}.
 
\subsubsection{Robustness Under DVL Sensor Fault}
\label{sec:dvl_fault}
 
\begin{figure}[!ht]
  \centering
  \includegraphics[width=0.85\linewidth]{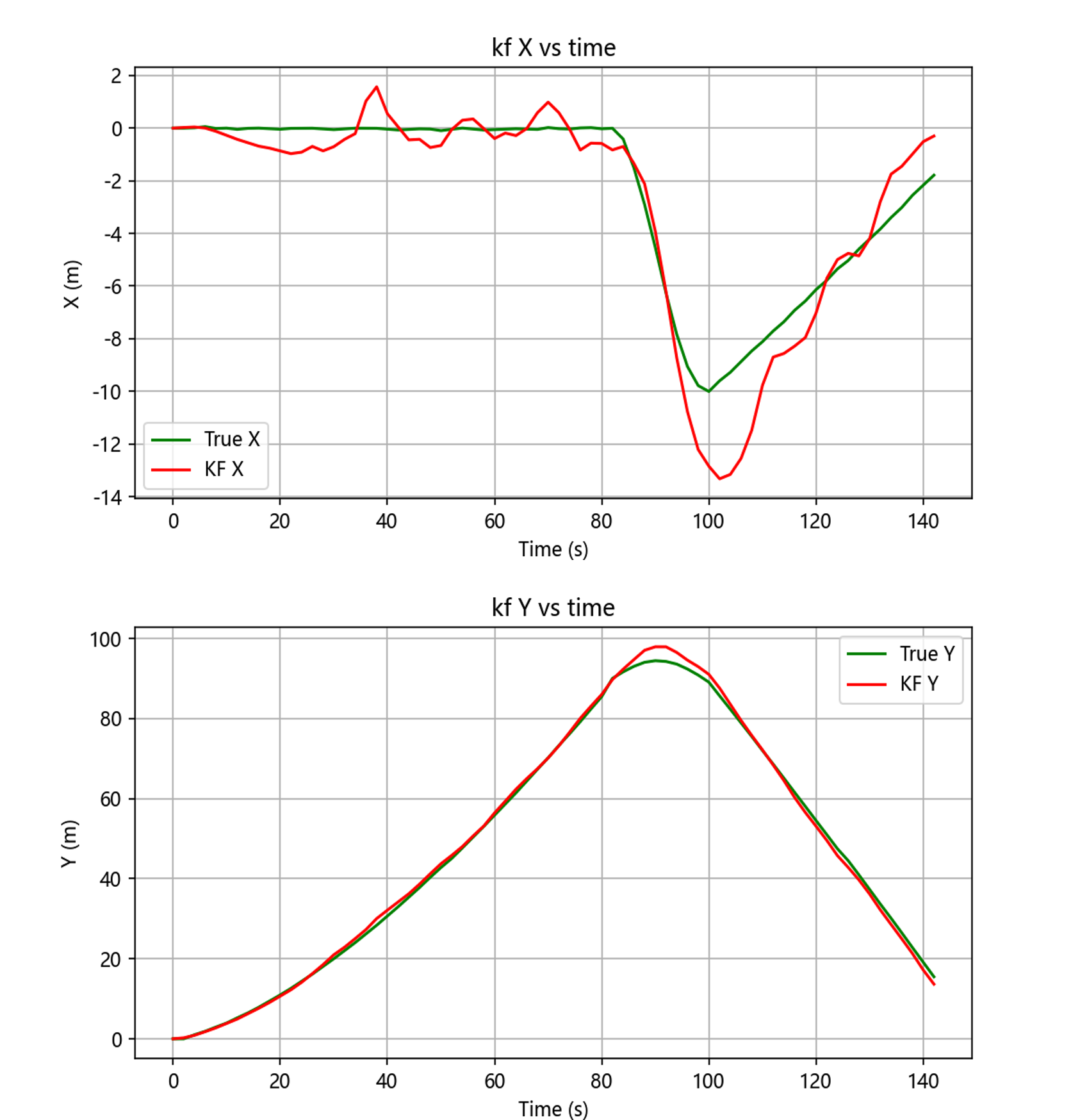}\\[4pt]
  \includegraphics[width=0.85\linewidth]{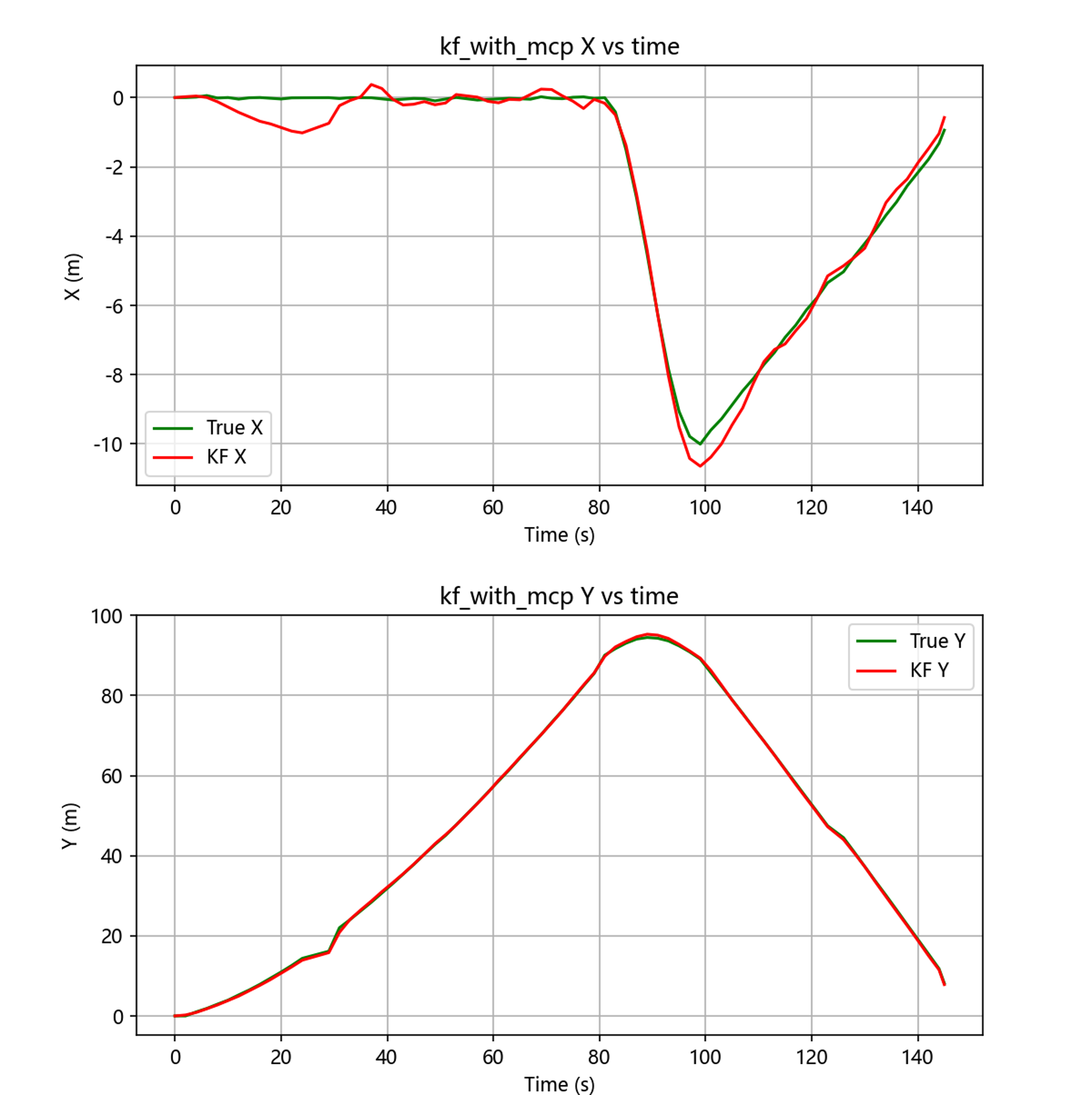}
  \caption{Lateral ($X$) and along-track ($Y$) time histories under the DVL lateral-bias fault, expressed in a planar frame aligned with the initial heading. \textit{Top}: \texttt{kf\_only}, with fixed KF covariances; the lateral channel deviates noticeably from the ground truth during and after the turn. \textit{Bottom}: \texttt{kf\_mcp}, where the LLM rescales the KF covariances at fault detection; the lateral residual is reduced and the along-track channel remains well tracked. Green: ground truth; red: KF estimate.}
  \label{fig:dvl_xy}
\end{figure}

\begin{figure}[!ht]
  \centering
  \includegraphics[width=\linewidth]{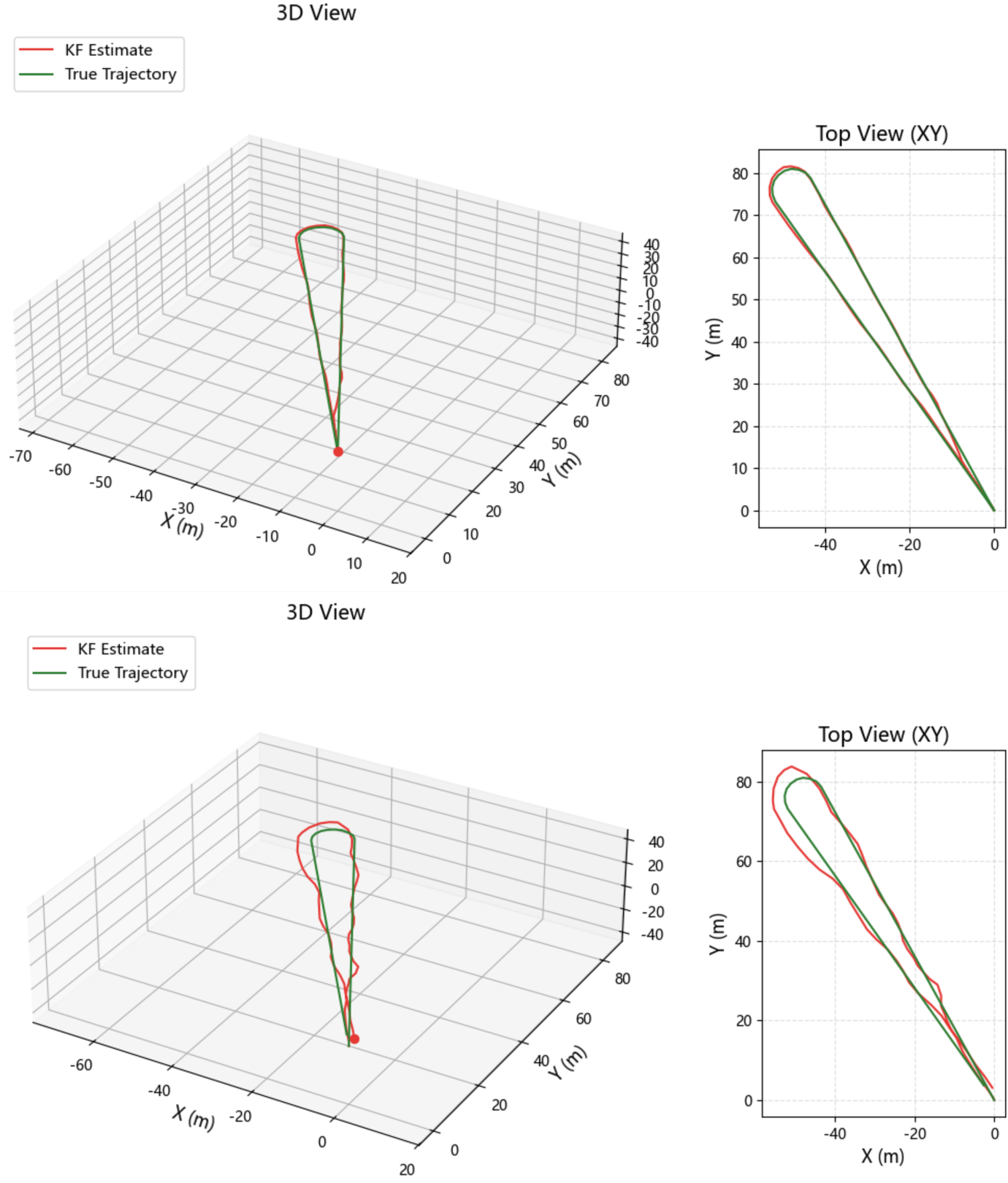}
  \caption{Trajectory-level comparison under the DVL lateral-bias fault. \textit{Top row}: \texttt{kf\_mcp}, with the LLM-issued MCP covariance update; the KF estimate (red) overlays the ground truth (green) throughout the loop. \textit{Bottom row}: \texttt{kf\_only}, with fixed covariances; the KF estimate exhibits visible lateral inflation and an outward bow on the return leg. Left column: 3D view; right column: top-down $XY$ view.}
  \label{fig:dvl_phases}
\end{figure}

Beyond the actuator-fault and environmental-disturbance scenarios already presented, the LASSA framework was evaluated under a sensor-degradation fault to assess its robustness against onboard state-estimation errors. The UUV's planar position is fused from DVL and GPS measurements through a Kalman filter (KF) at a sampling period of $\Delta t = 2.0$~s; at step $k = 15$, a persistent DVL lateral-velocity bias of $0.8$~m/s (with noise $\sigma = 0.5$~m/s) is injected. Two groups are compared on identical fault traces: \texttt{kf\_only}, with fixed measurement covariances; and \texttt{kf\_mcp}, in which the LLM issues an MCP tool call at fault detection that rescales the covariances $(Q_\mathrm{vel}, R_\mathrm{GPS}, R_\mathrm{DVL})$ by $(\times 8.0, \times 0.5, \times 20.0)$, downweighting the corrupted DVL channel and elevating the GPS contribution. The time-domain comparison is shown in Figure~\ref{fig:dvl_xy} and the spatial comparison in Figure~\ref{fig:dvl_phases}.
 
In the \texttt{kf\_only} group (Figure~\ref{fig:dvl_xy}, top), once the lateral DVL bias activates, the KF lateral-channel estimate progressively departs from the ground truth, and a residual offset remains visible through the turn segment; the along-track channel exhibits a smaller residual since the injected bias is primarily lateral. With fixed covariances, the KF treats the biased DVL signal as a credible velocity observation and propagates the error into the position estimate. In the \texttt{kf\_mcp} group (Figure~\ref{fig:dvl_xy}, bottom), the lateral residual is markedly reduced, and Figure~\ref{fig:dvl_phases} shows at the spatial level that the KF estimate overlays the ground-truth trajectory closely across the entire loop, while the along-track channel remains well tracked. The rescaled covariances effectively suppress the corrupted DVL channel and let the GPS observation dominate the fused estimate during the fault window, without modifying the KF structure.
 
This experiment demonstrates that the LASSA pipeline can extend beyond actuator and environmental faults to handle onboard sensor-degradation faults. Notably, the LLM selected and parameterised a sensor-fusion reconfiguration, a recovery action of a different category (covariance reweighting rather than geometric replanning) than the preceding scenarios---in response to a fault that was never explicitly enumerated in the design. Here MCP denotes the Model Context Protocol; the proposed covariance scaling factors lie outside the geometric strategy schema $\Theta(t)$ of Section~\ref{sec:llm}, and are admitted only after a parameter-bound check (positive-definiteness preservation and bounded scaling range) before being applied to the KF.
 
\subsubsection{Quantitative Summary}
 
Exp-S comprises three sub-scenarios with partly disjoint parameter sets, which are summarised separately. Key metrics for the steering-lock fault, cross-current disturbance, and DVL sensor-fault scenarios are listed in Tables~\ref{tab:sim_results}, \ref{tab:env_results}, and \ref{tab:dvl_results}, respectively. Under the steering-lock fault, the LASSA closed-loop pipeline operates correctly under continuous measurement noise: deviation detection, LLM-driven replanning, solver verification, and waypoint-tracking execution cooperate seamlessly, and the vehicle reaches the designated recovery point despite the transient heading-lock constraint imposed during the fault-response phase. Under the cross-current disturbance, the LLM autonomously inferred a rudder-bias correction of $\approx -11.5^\circ$ and replanned a straight-line path to the mission target, allowing the UUV to overcome the persistent lateral force. Under the DVL sensor-degradation fault, the LLM-issued MCP covariance update reduces the peak lateral-channel transient from $\approx 3.3$~m to $\approx 0.8$~m, a $\approx 76\%$ reduction relative to the fixed-covariance baseline.
 
\begin{table}[!ht]
  \centering
  \caption{Key metrics for the steering-lock fault scenario in Exp-S}
  \label{tab:sim_results}
  \begin{tabular}{ll}
    \toprule
    Metric & Value \\
    \midrule
    Nominal speed                      & 2.0~m/s \\
    State report period                & 2.0~s \\
    Cross-track deviation threshold    & 10~m \\
    Initial turning radius             & 5~m \\
    Replanned turning radius           & 10~m \\
    Replanned waypoint speed           & 1.0~m/s \\
    Steering-unlock depth delta        & 0.2~m \\
    Deviation detected                 & Yes \\
    Replanning triggered               & Yes \\
    Steering-lock phase executed       & Yes \\
    Mission completed                  & Yes \\
    \bottomrule
  \end{tabular}
\end{table}
 
Representative screenshots of the simulation monitoring interface at the initial-navigation stage and at the fault-triggered replanning stage of the steering-lock scenario are shown in Figure~\ref{fig:sim_ui_normal} and Figure~\ref{fig:sim_ui_fault}, respectively. The interface panels display the real-time FSM state, solver deviation reports, and the LLM I/O log, providing a transparent view of the agent's decision process at each stage.
 
\begin{center}
  \includegraphics[width=0.95\linewidth]{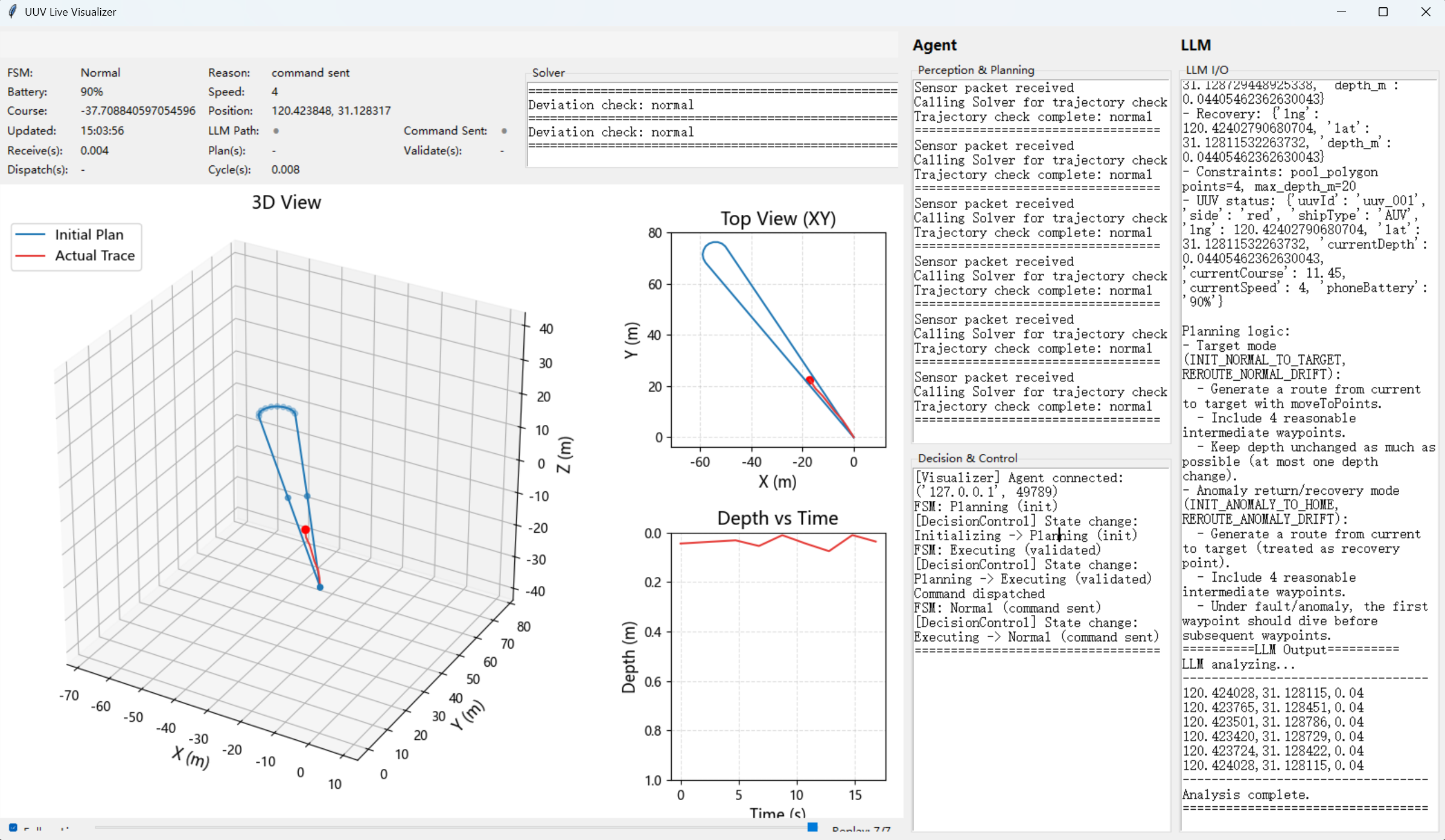}
  \captionof{figure}{Simulation monitoring interface during the initial navigation stage: the Solver panel reports ``Deviation check: normal'' at each cycle; the LLM I/O panel shows the planned waypoint coordinates output by the LLM; the FSM transitions through Planning~(init) $\rightarrow$ Executing~(validated) $\rightarrow$ Normal~(command~sent)}
  \label{fig:sim_ui_normal}
\end{center}
 
\begin{center}
  \includegraphics[width=0.95\linewidth]{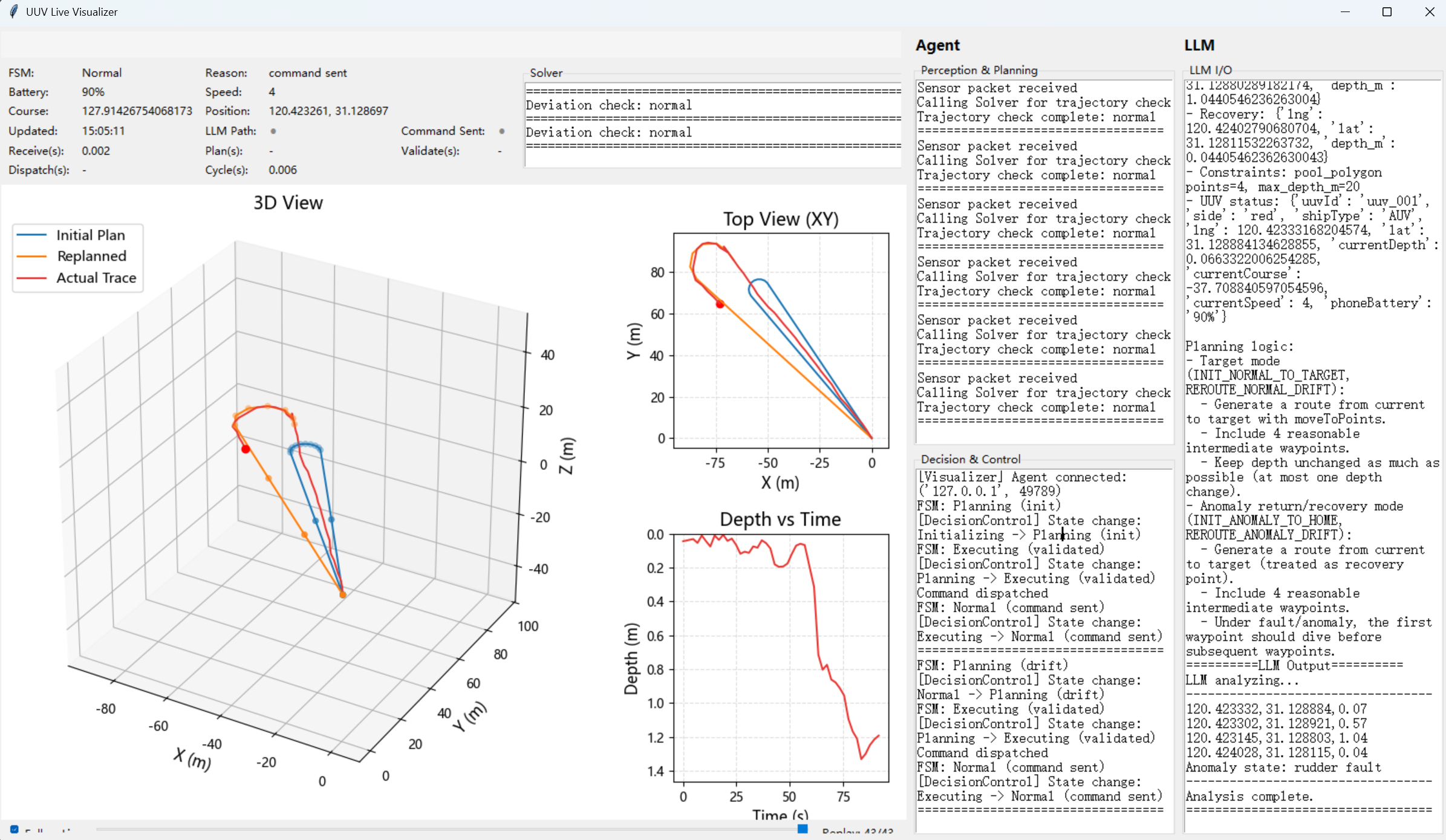}
  \captionof{figure}{Simulation monitoring interface during the fault-triggered replanning stage: the Decision \& Control panel shows the FSM transitioning to Planning~(drift); the LLM output panel explicitly identifies ``Anomaly state: rudder fault'' and generates a dive-then-return waypoint sequence; the depth profile shows the vehicle diving to approximately 1.4~m during the steering-lock phase before the replanned return is executed}
  \label{fig:sim_ui_fault}
\end{center}
 
\begin{table}[!ht]
  \centering
  \caption{Key metrics for the cross-current disturbance scenario in Exp-S}
  \label{tab:env_results}
  \begin{tabular}{ll}
    \toprule
    Metric & Value \\
    \midrule
    Cross-current lateral velocity            & 0.3~m/s \\
    Cross-current direction                   & $90^\circ$ (heading) \\
    Cross-track deviation threshold           & 10~m \\
    Time to deviation trigger                 & $\approx 34$~s ($\approx 17$ steps) \\
    LLM-inferred rudder-bias correction       & $\approx -11.5^\circ$ \\
    Replanned waypoint speed                  & 1.0~m/s \\
    Mission target reached                    & Yes \\
    \bottomrule
  \end{tabular}
\end{table}
 
\begin{table}[!ht]
  \centering
  \caption{Key metrics for the DVL sensor-fault scenario in Exp-S}
  \label{tab:dvl_results}
  \begin{tabular}{ll}
    \toprule
    Metric & Value \\
    \midrule
    DVL lateral-velocity bias                      & 0.8~m/s \\
    Fault trigger step                             & $k = 15$ \\
    Sampling period $\Delta t$                     & 2.0~s \\
    MCP covariance scaling $(Q_\mathrm{vel},R_\mathrm{GPS},R_\mathrm{DVL})$ & $( 8.0,\, 0.5,\, 20.0)$ \\
    Peak lateral deviation, \texttt{kf\_only}      & $\approx 3.3$~m \\
    Peak lateral deviation, \texttt{kf\_mcp}       & $\approx 0.8$~m \\
    Peak transient reduction                       & $\approx 76\%$ \\
    \bottomrule
  \end{tabular}
\end{table}
 
Exp-S complements the lake experiments by demonstrating that the LASSA strategy pipeline is robust to continuous sensor drift and can execute a multi-phase fault recovery encompassing deviation detection, cognitive replanning, physical verification, and waypoint tracking, entirely within the closed-loop architecture and without any human intervention or remote operator input. The three Exp-S sub-scenarios collectively exercise failure modes that are structurally distinct from the lake experiments: whereas Exp-F tests yaw-authority degradation under a physically removed rudder, the steering-lock sub-scenario tests the ability to sequence a conditional dive-then-turn recovery that introduces a temporal dependency between fault response and actuation capability, the cross-current sub-scenario tests robustness against a persistent environmental disturbance, and the DVL sub-scenario tests recovery from an onboard state-estimation degradation; together they exercise three different failure modes of the same replanning pipeline.

\subsection{LLM Ablation Study}
\label{sec:ablation_results}

\subsubsection{Model Comparison under a Fixed Prompt}

\begin{figure}[!ht]
  \centering
  \includegraphics[width=\linewidth]{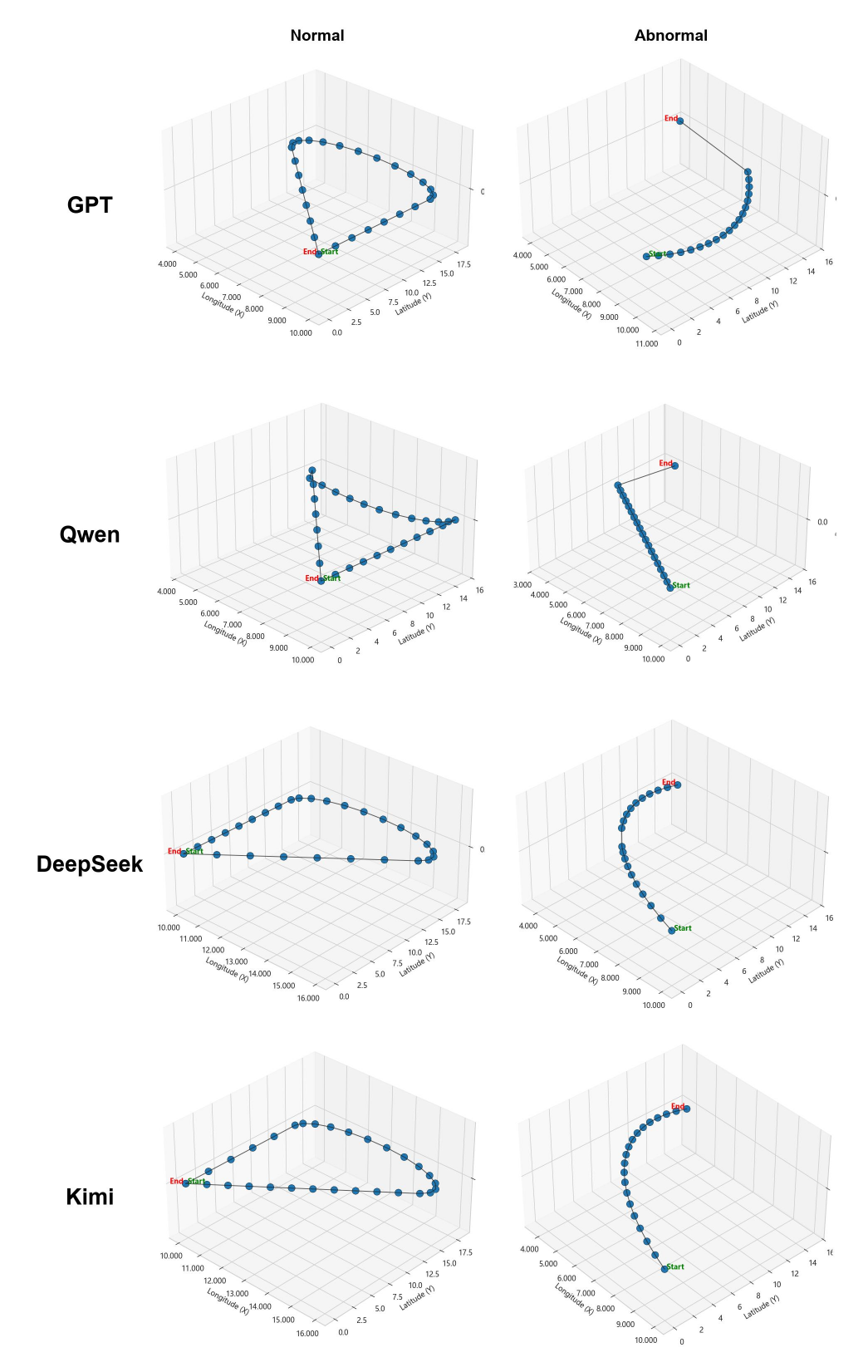}
  \caption{Model comparison. Each row shows one LLM backbone (GPT-5.3, Qwen3-max, DeepSeek-V3.2, Kimi~K2.5 from top to bottom); left column: normal scenario; right column: abnormal scenario. Kimi~K2.5 achieves the highest geometric fidelity in the normal scenario and the most consistent constraint-conflict response in the abnormal scenario; Qwen3-max exhibits the most severe deviation under the infeasible constraint.}
  \label{fig:ablation_g1}
\end{figure}

The planned paths for all four models under both scenarios are shown in Figure~\ref{fig:ablation_g1}, where each row corresponds to one model (GPT-5.3, Qwen3-max, DeepSeek-V3.2, Kimi~K2.5 from top to bottom) and the two columns show the normal and abnormal scenarios respectively.

\textbf{Normal scenario.} All four models generated closed navigation paths with arc segments (left column of Figure~\ref{fig:ablation_g1}). Kimi~K2.5 produced a smooth closed loop with uniform sampling, accurate arc tangency at both endpoints, and full boundary compliance, achieving the highest geometric fidelity among the four. DeepSeek-V3.2 produced a similarly smooth closed loop that closely matches the geometric target. GPT-5.3 generated a valid closed path with a slightly different boundary fit. Qwen3-max produced a broadly valid path but exhibited a notable kink at the arc transition, indicating reduced tangency accuracy compared with Kimi~K2.5 and DeepSeek-V3.2.

\textbf{Abnormal scenario.} The performance divergence across models was markedly more pronounced under the contradictory arc-radius constraint (right column of Figure~\ref{fig:ablation_g1}). Kimi~K2.5 recognised the constraint conflict and returned a best-effort partial arc with a reduced feasible radius, producing a geometrically consistent approximation rather than a random invalid path, demonstrating the strongest constraint-conflict reasoning among the four. DeepSeek-V3.2 produced a C-shaped partial arc that represents a plausible sub-arc approximation. GPT-5.3 generated a partial arc connected by a straight segment, indicating incomplete arc generation under the infeasible constraint. Qwen3-max produced a near-straight two-point path, the most severe geometric deviation among the four, suggesting a fundamental failure to process the contradictory constraint.

\subsubsection{Prompt Variant Comparison under Kimi K2.5}

\begin{figure}[!ht]
  \centering
  \includegraphics[width=\linewidth]{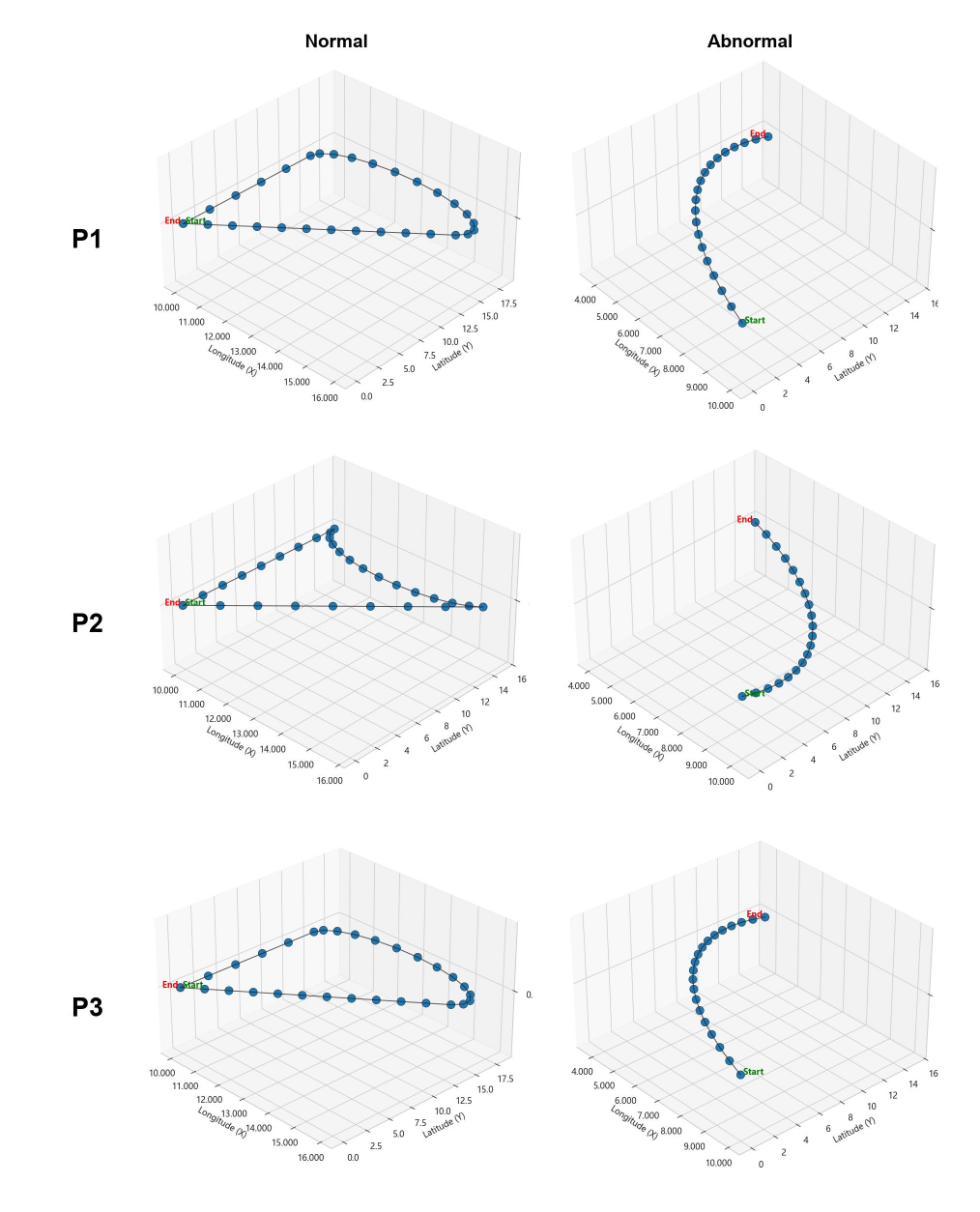}
  \caption{Prompt variant comparison using Kimi~K2.5. Rows correspond to the detailed (P1), standard (P2), and minimal (P3) prompt variants from top to bottom; left column: normal scenario; right column: abnormal scenario. The detailed variant (P1) achieves the highest arc quality in the normal scenario and the most geometrically consistent constraint-conflict response in the abnormal scenario; the minimal variant (P3) shows the greatest quality degradation under both conditions.}
  \label{fig:ablation_g2}
\end{figure}

The planned paths for the three prompt variants under both scenarios are shown in Figure~\ref{fig:ablation_g2}, where rows correspond to the detailed (P1), standard (P2), and minimal (P3) variants from top to bottom.

\textbf{Normal scenario.} The three prompt variants produced visibly different path qualities (left column of Figure~\ref{fig:ablation_g2}). The detailed variant (P1) produced a smooth closed loop with uniform sampling and accurate tangency, confirming that this configuration reliably extracts maximum planning accuracy from the model. The standard variant (P2) also produced a smooth closed loop that is geometrically valid, though with marginally less precise tangency at the arc start. The minimal variant (P3) produced a path with a distinctive bump at the arc apex and uneven sampling density; while the boundary geometry is met, the arc quality is noticeably lower, reflecting the model's difficulty in inferring implicit constraints from an underspecified prompt.

\textbf{Abnormal scenario.} Prompt precision had a decisive effect on constraint-conflict handling (right column of Figure~\ref{fig:ablation_g2}). The detailed variant (P1) produced a compact best-effort partial arc with the infeasible radius implicitly reduced to a feasible approximation. The standard variant (P2) generated a C-shaped partial arc that is geometrically plausible but does not explicitly flag the infeasibility. The minimal variant (P3) produced a smooth quarter-arc that terminates without closing. This result appears locally valid but fails to account for the global path-closure constraint violated by the infeasible radius, indicating that the underspecified prompt caused the model to solve a simpler sub-problem rather than detecting the full constraint conflict.

\subsubsection{Summary and Deployment Implications}

The ablation results yield two practical conclusions. First, \textbf{Kimi~K2.5} is selected as the LLM backbone for the lake and simulation experiments: it consistently outperformed the other candidates in both constraint satisfaction under normal conditions and constraint-conflict reasoning under adversarial conditions, demonstrating the strongest combination of geometric planning accuracy and infeasibility recognition. Second, the \textbf{detailed prompt variant} is adopted as the standard template: the structured geometric description, explicit tangency statements, per-segment sampling allocation, and zero-redundancy output format collectively improve path quality and reduce hallucination of infeasible trajectories to a negligible level.

These selections are consistent with the broader finding that LLM output quality in constrained trajectory planning depends critically on both model capability and prompt design: model capability determines the ceiling of achievable planning accuracy, while prompt precision determines how reliably that ceiling is reached in practice. The LASSA architecture benefits from both factors, as the prompt engine (Section~\ref{sec:llm}) is specifically designed to provide structured, constraint-rich context to the LLM at each replanning invocation, and the solver layer provides a hard safety net that prevents any remaining geometric errors from reaching the actuators.

\section{Conclusion}

This paper has presented the LASSA architecture for fault-state UUV autonomous navigation under communication-constrained conditions, integrating LLM-based cognitive reasoning, agent scheduling, and solver-based physical verification into a unified closed-loop pipeline requiring no real-time remote intervention.

LASSA demonstrates four unique and irreplaceable advantages over existing control methods. First, \textbf{autonomous reasoning without hard-coding}: the LLM generates adaptive replanning strategies by reasoning over vehicle state and fault context, avoiding rule-combination explosion and handling unanticipated fault configurations without reprogramming, which improves generalisation to complex and compound fault scenarios. Second, \textbf{hallucination suppression}: the solver certifies every LLM-generated trajectory against physical constraints before actuator dispatch, preventing infeasible outputs from reaching the vehicle. Third, \textbf{interpretability}: the full decision path from abnormality detection through strategy generation to solver certification is explicit and auditable, a property absent from end-to-end learned controllers. Fourth, \textbf{architectural completeness}: the LASSA architecture employs a fast–slow dual-loop time-sharing control structure. It deeply integrates sensor perception, autonomous reasoning, task scheduling, solver verification, and actuator control into an integrated end-to-end closed-loop framework. While substantially enhancing the autonomy of the control system, this architecture meets the high-precision and real-time performance requirements of low-level control.

Lake and simulation experiments validated the framework end-to-end, confirming fault detection within a five-cycle sliding window, first-attempt solver certification, and successful mission completion under degraded actuator authority with no false alarms. The current work covers four fault scenarios---lower-rudder failure, cross-current disturbance, and DVL sensor degradation, plus the steering-lock variant---in a controlled environment; future work will address compound and concurrent fault scenarios, onboard LLM deployment, and dynamic boundary updates for open-water operation.
\bibliographystyle{unsrt}
\bibliography{related_works_refs}

\end{document}